\documentclass[preprint,review,12pt]{elsarticle}

\usepackage[utf8]{inputenc}
\usepackage{graphicx}
\usepackage{amssymb}
\usepackage{amsmath}
\usepackage{amsfonts}
\usepackage{tabularx,colortbl}
\usepackage{multirow}
\usepackage{color}
\usepackage[hidelinks]{hyperref}
\usepackage[english]{babel}
\usepackage{algorithm}
\usepackage{algorithmic}
\usepackage[font={footnotesize}]{caption}
\usepackage{subcaption}
\usepackage[title]{appendix}

\definecolor{Orange}{rgb}{1,0.64,0}
\definecolor{lgray}{rgb}{0.92,0.92,0.92}

\newtheorem{theorem}{Theorem}[section]

\journal{Information Sciences}
\makeindex

\begin{document}
\begin{frontmatter}

\title{An Agent-Based Algorithm exploiting Multiple Local Dissimilarities for Clusters Mining and Knowledge Discovery}

\author[address1]{{Filippo Maria} Bianchi\corref{cor1}}
\ead{filippo.binachi@ryerson.ca}
\cortext[cor1]{Corresponding Author}
\author[address1]{Enrico Maiorino}
\ead{enrico.maiorino@uniroma1.it}
\author[address2]{Lorenzo Livi}
\ead{llivi@scs.ryerson.ca}
\ead[url]{https://sites.google.com/site/lorenzlivi/}
\author[address1]{Antonello Rizzi}
\ead{antonello.rizzi@uniroma1.it}
\ead[url]{http://infocom.uniroma1.it/~rizzi/}
\author[address2]{Alireza Sadeghian}
\ead{asadeghi@ryerson.ca}
\ead[url]{http://www.scs.ryerson.ca/~asadeghi/}

\address[address1]{Dept. of Information Engineering, Electronics, and Telecommunications, SAPIENZA University of Rome, Via Eudossiana 18, 00184 Rome, Italy}
\address[address2]{Dept. of Computer Science, Ryerson University, 350 Victoria Street, Toronto, ON M5B 2K3, Canada}

\begin{abstract}
We propose a multi-agent algorithm able to automatically discover relevant regularities in a given dataset, determining at the same time the set of configurations of the adopted parametric dissimilarity measure yielding compact and separated clusters.
Each agent operates independently by performing a Markovian random walk on a suitable weighted graph representation of the input dataset.
Such a weighted graph representation is induced by the specific parameter configuration of the dissimilarity measure adopted by the agent, which searches and takes decisions autonomously for one cluster at a time.
Results show that the algorithm is able to discover parameter configurations that yield a consistent and interpretable collection of clusters. Moreover, we demonstrate that our algorithm shows comparable performances with other similar state-of-the-art algorithms when facing specific clustering problems.
\end{abstract}
\end{frontmatter}

%\linenumbers

%%%%%%%%
\section{Introduction}
Finding characterizing regularities in data is an important knowledge discovery task, which can be exploited for a multitude of purposes. 
When there is not any a-priori knowledge on the dataset at hand, it could be useful to perform an initial analysis of the data in order to learn how to compare the elements in a meaningful way, so that relevant patterns in the dataset can be discovered.
Clustering \cite{kannan2004clusterings,6414624,Cao:2013:TIK:2461537.2461576,pedrycz2005knowledge,cagata,6378449,zhang2014interval} is a well-established approach that can be used to this end.
Among the many solutions available in this field, it is worth citing those clustering techniques based on graph-theoretical results and multi-agent systems \cite{10.1109/TPAMI.2012.226,ATabrizi20135772,Ferrer:2009:GKC:1617915.1617963,Galluccio:2012:GBK:2184924.2185067,Galluccio201396,azran2006spectral,north2014theoretical,agogino2006efficient,giannella2004multi}.
Graph-based techniques have the fundamental advantage of mapping the original problem onto a ``dimensionless'' object: the graph.
Moreover, graph theory offers a plateau of theoretical results to be exploited by effective algorithms, which easily integrate with the agent-based paradigm.
Typical settings involving the interplay of both approaches include random walk (RW) based algorithms \cite{5693955,gallesco2011note}, in which agents move and interact on the graph via specific (probabilistic) mechanisms.

When there is uncertainty about the nature of the dataset at hand, a fundamental issue is the definition of the dissimilarity among the input patterns \cite{odse,Schleif_simbad2013,Duin2012826}, since the specific dissimilarity measure adopted by the data mining procedure affects the possibility of discovering meaningful regularities.
Depending on the application at hand, data can be collected and represented relying on several different formalisms \cite{si_asoc_grc}.
Accordingly, many (non metric) parametric dissimilarity measures could be designed depending on the specific task.
Recently, there is a steady increasing interest in using several, possibly heterogeneous, dissimilarity measures at the same time \cite{kim_duin__multiplediss__2009,6399461,Queiroz20133383,Bereta20131213,gonen2011multiple}.
Regardless of the number of dissimilarity measures, the setting of their characterizing parameters is what really allows to discover the relevant information hidden in the data.

Metric learning \cite{6588959,4674367,yin2012semi,zhang2012semi,chang2012boosting} is an important subfield of pattern recognition. Techniques in this field deal with the problem of learning an optimal setting of the parameters characterizing the particular dissimilarity for the problem at hand -- usually it is assumed to be a metric distance.
For a given dissimilarity measure, it is possible to distinguish two main approaches \cite{Mu20132337}: those trying to determine a partition of data, and those that focus on searching for isolated clusters surrounded by uncategorized data. Local description of data is of particular interest, since it allows to characterize the input data by means of a heterogeneous collection of descriptions \cite{Bereta20131213}.

In this paper we propose the \textit{Local Dissimilarities - Agent Based Clusters Discoverer} (LD-ABCD) algorithm.
LD-ABCD is designed to discover (learn) configurations of a parametric dissimilarity measure yielding at least a well-formed cluster in the data.
Cluster discovery is implemented by means of multiple RWs that are performed independently by several agents on the graph representing the dataset.
Each agent first selects a specific parameter configuration (PC), with which it constructs a weighted graph representing the relations among the input patterns.
The behavior of a RW is thus dependent on the specific configuration of the parameters. During a RW, an agent searches and takes decisions autonomously for one cluster at a time. A suitable online mechanism is designed to decide whether a set of patterns found (i.e., ``walked upon'') by an agent should be accepted or rejected as a meaningful cluster.
To this end, we heavily exploit the graph conductance concept \cite{kannan2004clusterings}.
We demonstrate the validity of our approach by performing different types of experiments.
First, we compare LD-ABCD with respect to (w.r.t.) three different state-of-the-art graph-based clustering algorithms over suitable clustering problems. In particular, we evaluate the capability of the considered algorithms to discover clusters composed of patterns belonging to the same (predefined) class.
Successively, we evaluate the capability of LD-ABCD of discovering relevant PCs (RPCs), that is, those that yield well-formed clusters. Additionally, we provide demonstrative examples introducing the concept of equivalency among PCs.
Finally, we provide a comparison between two variants of the LD-ABCD algorithm.

The remainder of the paper is structured as follows.
In Sec. \ref{sec:algodesc} we introduce LD-ABCD, describing in detail all relevant stages of the algorithm.
In Sec. \ref{sec:exploration_exploitation} we present a variant of LD-ABCD that exploits two diverse families of agents.
Experimental evaluations are presented and discussed in Sec. \ref{sec:exps}, while in Sec. \ref{sec:conclusions} we show our conclusions. Finally, Appendix \ref{sec:graph_conductance} provides the technical details related to the definition and calculation of the graph conductance.

\subsection{Related Works}
The work that we present in this paper is related to several different topics, specifically graph clustering, conductance evaluation, metric learning, and agent-based computing. At the best of the author knowledge, it was not possible to identify other works that treat the problem of clustering and knowledge discovery with approaches similar to the one that we proposed.
The aim of this section is helping the reader to contextualize our work and to correctly identify the concepts to which our work is related.

In particular LD-ABCD identifies clusters on a dataset that is represented through a labeled graph: graph clustering is a well-known problem and it has been addressed in many other works \cite{10.1109/TPAMI.2012.226,ATabrizi20135772,Ferrer:2009:GKC:1617915.1617963,Galluccio:2012:GBK:2184924.2185067,Galluccio201396,north2014theoretical}. 
Such clusters are discovered by different agents, which operate according to a paradigm inspired by the multi-agent systems that can be found in the literature \cite{azran2006spectral, Negenborn:10h, 6048999, Chaimontree_ma__2012,5400314, 5693955, north2014theoretical}. Each agent examines the patterns by performing a RW \cite{5693955,gallesco2011note} on the graph that represents the dataset and tries to group them in different clusters. Once the clusters are identified, they are evaluated using the well-known conductance measurement \cite{kannan2004clusterings}, which is computed using numerical approximation techniques \cite{arora2008geometry,hoory2006expander,trefethen1997numerical,Leighton:1999:MMM:331524.331526, arora2009expander, madry2010fast, gkantsidis2003conductance, sarma2011estimating}.
Finally, each agent searches the clusters in the dataset using different configurations of the adopted dissimilarity measure, seeking for the ones that better characterize the set of elements contained in the cluster.
This procedure is strongly related to the problem of the metric-learning \cite{6588959,4674367,yin2012semi,zhang2012semi,chang2012boosting}, which is the task of determining the optimal parameters of a given metric distance.
However, in our case we make no a-priori assumptions on the adopted distance (which we call dissimilarity measure in our study).

%%%%%%%%
\section{The Proposed LD-ABCD Algorithm}
\label{sec:algodesc}
The proposed multi-agent algorithm is designed to operate over a general input domain, $\mathcal{X}$, which may not necessarily be a subset of $\mathbb{R}^n$.
Let $d:\mathcal{X}\times\mathcal{X}\rightarrow\mathbb{R}^+$ be a symmetric dissimilarity measure that depends on some parameters/weights, i.e., PCs, which we denote as $m\in\mathcal{M}$.
The main goal of the proposed algorithm is to determine all RPCs which are capable of inducing a well-formed cluster structure.
In this sense, our algorithm should be intended also as a ``knowledge discovery'' algorithm, since, in addition to the clusters discovered using local configurations of $d(\cdot, \cdot; m)$, it outputs all relevant settings of the parameters characterizing the dissimilarity measure, which may be useful in terms of interpretability of the data and related clusters.
Without loss of generality, we also assume that $\mathcal{M}=[0, 1]^D$, where $D$ is the number of parameters/weights characterizing $d(\cdot, \cdot; m)$.

\begin{figure}[ht!]
 \centering
 \includegraphics[viewport=0 0 917 519,scale=0.4,keepaspectratio=true]{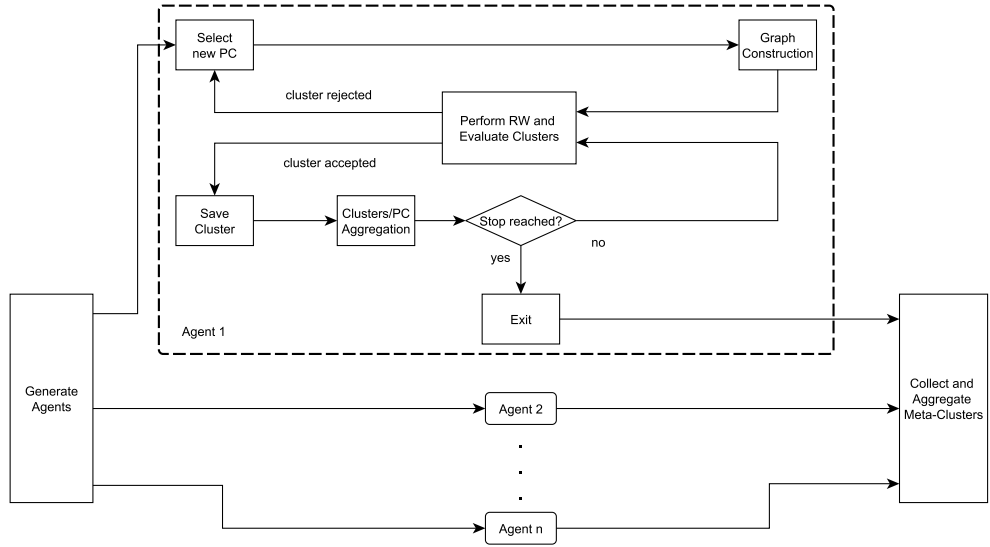}
 \caption{Overall schema of the LD-ABCD algorithm. Initially, several agents are generated and each of them performs a series of tasks, which are shown in detail for the first agent. At first a new PC is selected which is used for building a weighted graph of the dataset $\mathcal{S}$ and then a RW is performed on it. The set of visited nodes is treated as a cluster, which is evaluated by the agent. If the cluster is accepted it will be saved along with the other clusters found so far, otherwise if it is rejected the agent restarts the whole procedure from the selection of a new PC. Similar clusters are aggregated and the procedure goes on until the stop criterion has been met. When all the agents terminate their search, the found aggregated clusters are collected by a central unit which performs a final merging of similar solutions found by different agents.}
 \label{fig:schema1}
\end{figure}

Fig. \ref{fig:schema1} provides the overall high-level schema of the LD-ABCD algorithm, together with details of the operations performed by a single agent within the proposed system.
Each agent $a_i$ uses a different PC $m_{j}^{(i)}$ for evaluating the dissimilarity among the patterns in the input dataset $\mathcal{S}=\{x_1, x_2, ..., x_n\}\subset\mathcal{X}$. The dataset is initially represented as a weighted complete undirected graph, $G_j=(\mathcal{V}, \mathcal{E}, w)$, where each edge $e_{kl}\in\mathcal{E}$ is characterized by a weight, $w(e_{kl}; m_j^{(i)})\in[0, 1]$, which depends on the dissimilarity $d(x_k, x_l; m_j^{(i)})$ evaluated with the specific $m_j^{(i)}$.
Each agent performs a Markovian RW \cite{Lovasz1996} on the graph $G_j$, visiting a number of vertices (nodes) until a quantity called ``energy'' is not depleted. The RW transition probabilities from one node to another are determined by the weight values of the edges between pairs of nodes and, hence, depend on the parameter configuration associated with the agent. When an agent $a_i$ equipped with the PC $m_j^{(i)}$ runs out of energy, the vertices visited so far during the RW are interpreted as the cluster $c_{h}(m_j^{(i)})\subset\mathcal{V}$ (or $c_{hj}$ for notation simplicity) found by the agent -- which corresponds also to the subgraph $g_{hj}$.
Therefore, each agent generates a single cluster at a time that is readily evaluated by the agent itself, which takes an autonomous decision on its acceptance. Since each agent generates the clusters independently from the others, the clusters retrieved by LD-ABCD may overlap (i.e. a given pattern can belong to more than one cluster) and, also, their union could not be equal to $\mathcal{V}$; thus LD-ABCD does not generate a partition of the data  (i.e. not all the patterns in the data set will belong to a cluster).
During its lifetime, an agent performs several RWs on different versions of the same graph, which depend on the adopted PCs. Since it is possible for an agent to find similar clusters when using different PCs, these are progressively aggregated in prototypes called \textit{meta-clusters}.
The algorithm proceeds as long as new distinct clusters are extracted or new PCs are discovered. When the stop criterion has been met the meta-clusters are returned along with the PCs associated to the set of clusters represented by each meta-cluster.
Finally, a centralized unit re-aggregates meta-clusters belonging to different agents according to their similarity to obtain new \textit{global} meta-clusters. The final solution returned by LD-ABCD is the collection of all global meta-clusters and their corresponding sets of associated PCs.
In LD-ABCD, the number of agents is defined a priori by the user and it remains the same during the execution. This number is supposed to be proportional to the available computational resources.

In the following, we provide the details about the tasks performed by a single agent during its lifetime. First, we discuss how the weighted graph is constructed over the input dataset (\ref{sec:graph_construction}). Then we focus on the implementation of the RWs (\ref{sec:random_walk_cluster}) and the evaluation of the discovered clusters (\ref{sec:clusterquality}). The procedure for managing the energy of the agents is discussed in Sec. \ref{sec:energyupdate}. In Sec. \ref{sec:new_metrics}, we describe the process of selection of the new PCs to be exploited, while in Sec. \ref{sec:cluster_aggr} we discuss the aggregation of the solutions found by different agents and the global convergence criterion of LD-ABCD (Sec. \ref{sec:mamld_convergence}). Finally we analyze the computational complexity of the algorithm in Sec. \ref{sec:compcomplexity}.

\subsection{Graph Construction}
\label{sec:graph_construction}
Let assume that an agent $a_i$ is equipped with the PC $m_j^{(i)}$, and let $\mathcal{S}$ be the dataset under analysis, with $n=|\mathcal{S}|$.
The corresponding weighted graph $G_j=(\mathcal{V}, \mathcal{E}, w)$, is described by the vertices $\mathcal{V}$, each one representing a pattern in $\mathcal{S}$, and by the edges $\mathcal{E}$, which are weighted by implementing $w(\cdot)$ as the exponential kernel:
\begin{equation}
\label{eq:edge_weight}
w(e_{lk}; m_j^{(i)}) = \text{exp}(- \tau_{\text{exp}} \cdot d(x_l, x_k; m_j^{(i)})).
\end{equation}

The setting of the parameter $\tau_{\text{exp}}\geq 0$ is an important issue and it will be discussed later in Sec \ref{sec:random_walk_cluster}.
A weighted graph can be described by the $n\times n$ weighted adjacency matrix $\mathbf{A}_j$, defined as:
\begin{equation}
\label{eq:wadj}
\mathbf{A}_j(l, k)=w(e_{lk}; m_j^{(i)}).
\end{equation}

Since the vertex set is not affected by the specific PC, we keep the related data in a shared data structure, which is accessible by all agents.
The edges, which instead can differ on the base of the specific PC, are stored ``locally'' by each agent, encoded in their weighted adjacency matrix.
The computational and space costs for $\mathbf{A}_j$ is quadratic in the number of vertices-patterns (such a matrix is always dense).
For large datasets, the construction of those graphs on a single computing machine could be unfeasible due to memory limitation.
By exploiting the distributed nature of the adopted agent-based modeling, we could easily elude this technical problem by suitably dispatching ``chunks'' of the original datasets among the various agents--machines.
This would imply a distributed communication mechanism that, at this stage of development of LD-ABCD, is not implemented yet.
Therefore, in the following we assume each agent to be able to process the input dataset as a whole.

\subsection{Random Walk for Cluster Search}
\label{sec:random_walk_cluster}
To perform a RW on $G_j$ we need to define the so-called transition matrix \cite{Lovasz1996}, $\mathbf{M}_j$, which is used by an agent to navigate among the vertices. $\mathbf{M}_j$ is defined as follows,
\begin{equation}
\label{eq:prob_matrix}
\mathbf{M}_j = \mathbf{D}_{j}^{-1} \mathbf{A}_j,
\end{equation}
where $\mathbf{D}_j$ is the (diagonal) degree matrix: $D_j(l,l) = \sum_{k=1}^{|\mathcal{V}|} A_j(l,k)$.
A RW can be effectively characterized by exploiting the stationary distribution (SD) $\boldsymbol\pi_j$ of the Markov process underlying the RW. The SD can be interpreted as the left eigenvector of $\mathbf{M}_j$, associated to the largest eigenvalue, i.e., 1.
Every complete and non bipartite graph has a stationary distribution \cite{Lovasz1996}, which can be conveniently defined by exploiting the so-called degree distribution,
\begin{equation}
\boldsymbol\pi_j(v_l) = \frac{D(l, l)}{2 |\mathcal{E}|}, \ \forall v_l\in \mathcal{V}.
\end{equation}

\begin{figure}
        \centering
        \begin{subfigure}[t]{0.4\textwidth}
                \includegraphics[width=\textwidth]{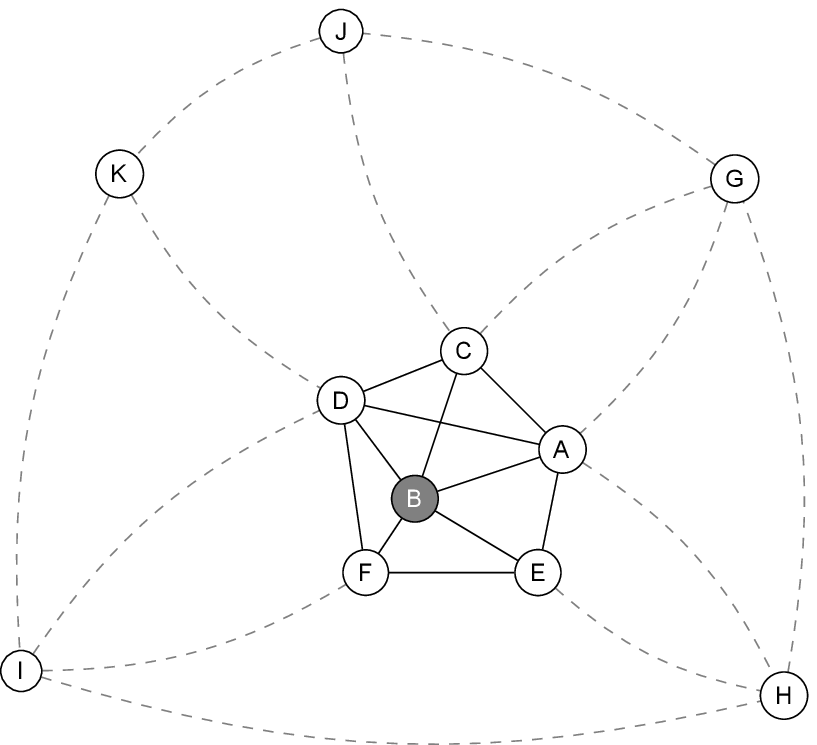}
                \caption{}
                \label{fig:grafico1}
        \end{subfigure}
        	\qquad
        ~ %add desired spacing between images, e. g. ~, \quad, \qquad, \hfill etc.
          %(or a blank line to force the subfigure onto a new line)
        \begin{subfigure}[t]{0.4\textwidth}
                \includegraphics[width=\textwidth]{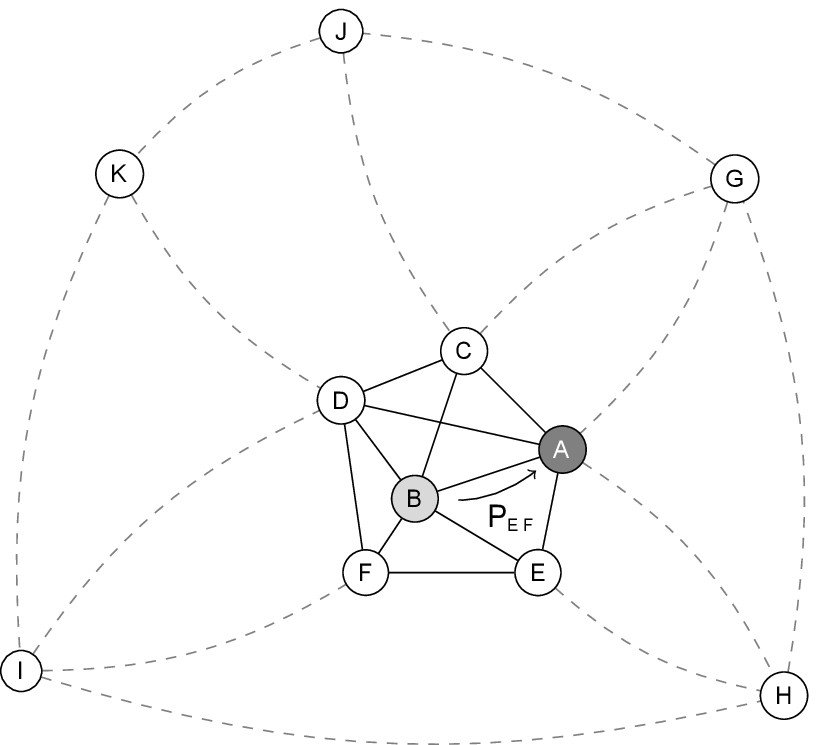}
                \caption{}
                \label{fig:grafico2}
        \end{subfigure}
        ~ %add desired spacing between images, e. g. ~, \quad, \qquad, \hfill etc.
          %(or a blank line to force the subfigure onto a new line)

        \begin{subfigure}[t]{0.4\textwidth}
                \includegraphics[width=\textwidth]{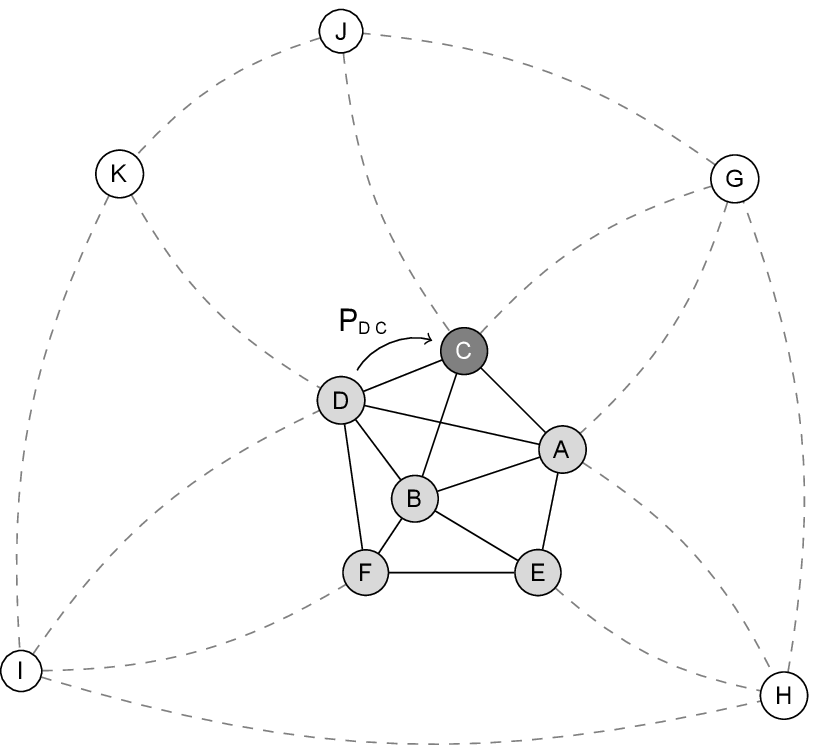}
                \caption{}
                \label{fig:grafico3}
        \end{subfigure}
        	\qquad
	~ %add desired spacing between images, e. g. ~, \quad, \qquad, \hfill etc.
          %(or a blank line to force the subfigure onto a new line)
        \begin{subfigure}[t]{0.4\textwidth}
                \includegraphics[width=\textwidth]{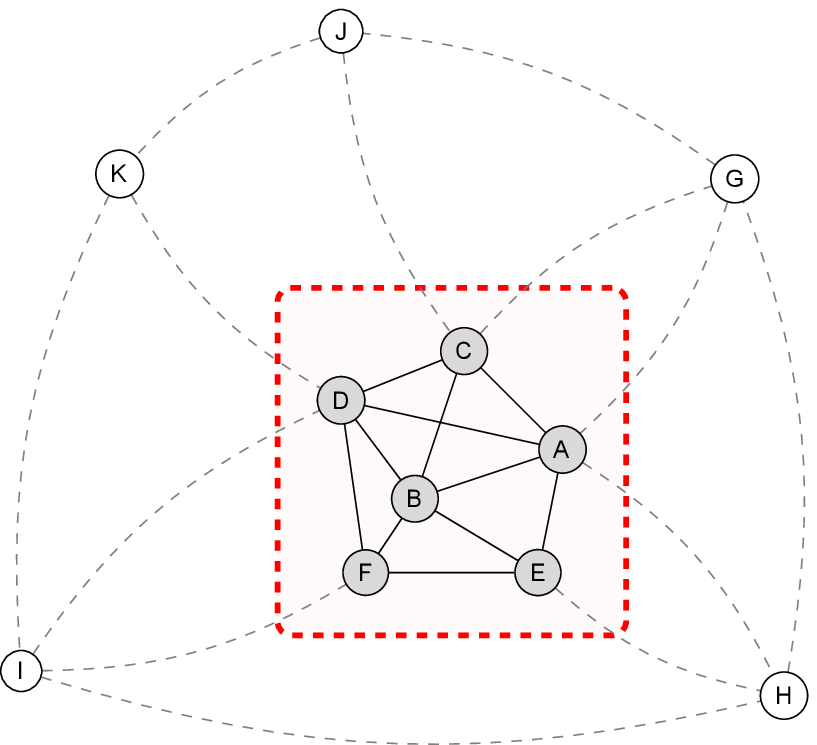}
                 \caption{}
                
                \label{fig:grafico4}
        \end{subfigure}
        \caption{RW example. Small dissimilarity values are represented with continuous lines, while the large ones are drew with dashed lines. The probability of moving from one node to another is given by the transition matrix of Eq. \ref{eq:prob_matrix}, which takes into account the magnitude of the dissimilarities. The current node is represented with dark gray color, while explored and unexplored nodes are represented, respectively, with light gray and white colors. The graph is fully connected but some edges are not shown for clarity. \textit{(a)} The agent starts from node B, which is a central node in the graph selected using the SD $\boldsymbol\pi$. \textit{(b)} The agent uses the transition matrix $\mathbf{M}$ for choosing the next node to visit. In this case, the node A is selected with probability $P_{\text{BA}}$ as the next node in the walk. \textit{(c)} The agent continues its walk moving from node D to node C. \textit{(d)} The resulting set of all the nodes visited by the agent during its walk.
}
        \label{fig:random_walk}
\end{figure}

We use the SD $\boldsymbol\pi_j$ for selecting the starting vertex $v_s$ from which an agent starts a RW, since highly central vertices will have higher probability according to the SD.
In this way, we let an agent start a RW from a dense region of the graph, rather than from a peripheral region in which it could be stuck or it could easily move to a vertex belonging to a ``different'' dense region (see Fig. \ref{fig:random_walk}).

A correct setting of $\tau_{\text{exp}}$ (\ref{eq:edge_weight}) is crucial, since it affects the behavior of the RW performed by an agent.
In fact, a higher value of $\tau_{\text{exp}}$ magnifies the edge weights between similar patterns, making less likely the unwanted transitions to vertices connected by low weights (i.e., dissimilar patterns).
Notably, if we assign to $\tau_{\text{exp}}$ a value that is too high, the lower weights could be excessively magnified. In this case, an agent would repeatedly move on a very small set of vertices, instead of exploring a larger portion of the graph (transition probabilities become degenerate).
On the other hand, assigning a too small value to $\tau_{\text{exp}}$ would lead to the opposite situation, as it would allow the agent to jump to different regions of the graph during the RW (transition probabilities become uniform).

In LD-ABCD, we heuristically set $\tau_{\text{exp}}$ as a value proportional to the average distance between the input patterns, evaluated using the dissimilarity measure configured with the PC currently selected by the agent,
\begin{equation}
 \tau_{\text{exp}} = \beta n^{-2}\sum_{l,k=1}^{n} \mathbf{A}_j(l,k),
\end{equation}
where $\beta$ is a user-defined value that is set empirically.

Of course, more accurate methods could be defined for estimating $\tau_{\text{exp}}$.
However, since in our setting $d(\cdot, \cdot)$ may be possibly not metric (and also not algebraic, i.e., which cannot be expressed in closed form), it is hard to find a strong relation among $\tau_{\text{exp}}$ and the transition probabilities.

\subsection{Cluster Quality Evaluation}
\label{sec:clusterquality}
An agent $a_i$ generates a cluster $c_{hj}$ during a RW performed on $G_j$ with the PC $m_j^{(i)}$, which consists in the set of vertices of the subgraph $g_{hj}$ visited during the RW (see Fig. \ref{fig:random_walk}).
In the following, we will refer equivalently to $g_{hj}$ and $c_{hj}$. Once a cluster $c_{hj}$ is returned by an agent $a_i$, the cluster can be either accepted or rejected, depending on its quality.
Intuitively, a cluster is considered to be good if it contains several elements, which are also very similar to each other according to the current PC.
A well-established measure used for evaluating the quality of a cluster associated to a subgraph of a larger graph is the conductance \cite{kannan2004clusterings}, $\phi(c_{hj})$, which quantifies how well knit is the subgraph internally and how many edges (with their associated weights) connected to vertices outside the cluster are cut.
In terms of clustering, a subgraph with low conductance represents a compact and populated cluster, which is also well-separated from the remaining elements of the dataset.
A straightforward method for evaluating the quality of a cluster then consists in defining a function $\text{CQ}_1$ directly proportional to its conductance:
\begin{equation}
 \label{eq:CQ1}
 \text{CQ}_1(c_{hj}) = 1 - \phi(c_{hj}).
\end{equation}

A cluster $c_{hj}$ is therefore accepted if $\text{CQ}_1(c_{hj}) \geq \tau_{\text{CQ}}$, where $\tau_{\text{CQ}}\geq 0$ is a user-defined threshold.

However, directly using the conductance as a quality measure of clusters discovered in different datasets could be not easily manageable. In fact, the value of the conductance of a cluster depends also on the configuration of the rest of the dataset and thus it could fall within very diverse ranges, making the decisions and interpretations regarding its quality a difficult task. Additionally, since in our work we made no assumption on the employed dissimilarity measure used for comparing the patterns, it is not easy to express in closed form the variation of the conductance as the values of the parameters of the dissimilarity value change.
Thus, given a dataset, it is hard to describe analytically the relation among the quality of the clusters and the used PCs.
For those reasons, we introduce here a new quantity for evaluating the quality of a cluster, which takes into account the properties of the whole graph constructed by using a specific PC. 
In particular, we assert that the quality of a cluster $c_{hj}$ is proportional to the closeness of its conductance, $\phi(c_{hj})$, to the minimum conductance of the whole graph $G_j$ (or simply the conductance of $G_j$), denoted as $\Phi(G_j)$.
The exact computation of $\Phi(G_j)$ is a NP-Hard problem \cite{kannan2004clusterings}, and hence it is not computationally feasible.
As a consequence, in this paper we use an approximation for $\Phi(G_j)$, defined through a pair of real numbers, $\mathrm{lb}(\Phi(G_j)), \mathrm{ub}(\Phi(G_j))$, which represent, respectively, the lower and the upper bound of the interval that contains the actual value of $\Phi(G_j)$.
These values can be computed exploiting the Cheeger’s inequality, by means of a procedure that is discussed in Appendix \ref{sec:graph_conductance}.
We introduce a novel cluster quality function, $\text{CQ}_2$, defined as:
\begin{equation}
\label{eq:CQ2}
\text{CQ}_2(c_{hj}) = 1 - \frac{\phi(c_{hj}) - \mathrm{lb}(\Phi(G_j))}{\mathrm{ub}(\Phi(G_j)) - \mathrm{lb}(\Phi(G_j))}.
\end{equation}

\begin{figure}[ht!]
 \centering
 \includegraphics[bb=0 0 1141 592,scale=0.35,keepaspectratio=true]{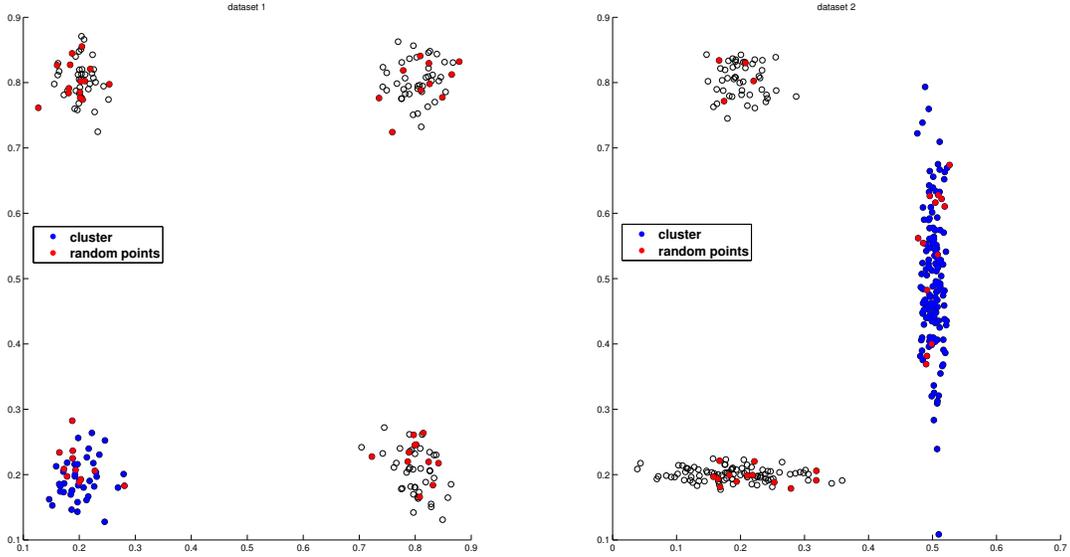}
 % datasetCQvsCOND.eps: 0x0 pixel, 300dpi, 0.00x0.00 cm, bb=0 0 1141 592
 \caption{Datasets considered to appreciate the differences in using $\text{CQ}_2$ rather than $\text{CQ}_1$. The blue dots represent a well-defined cluster, the red dots a bad cluster composed by randomly selected patterns, while the white dots represent the unselected patterns in the dataset.}
 \label{fig:ds_CQ}
\end{figure}

\begin{figure}[ht!]
 \centering
 \includegraphics[bb=0 0 992 598,scale=0.47,keepaspectratio=true]{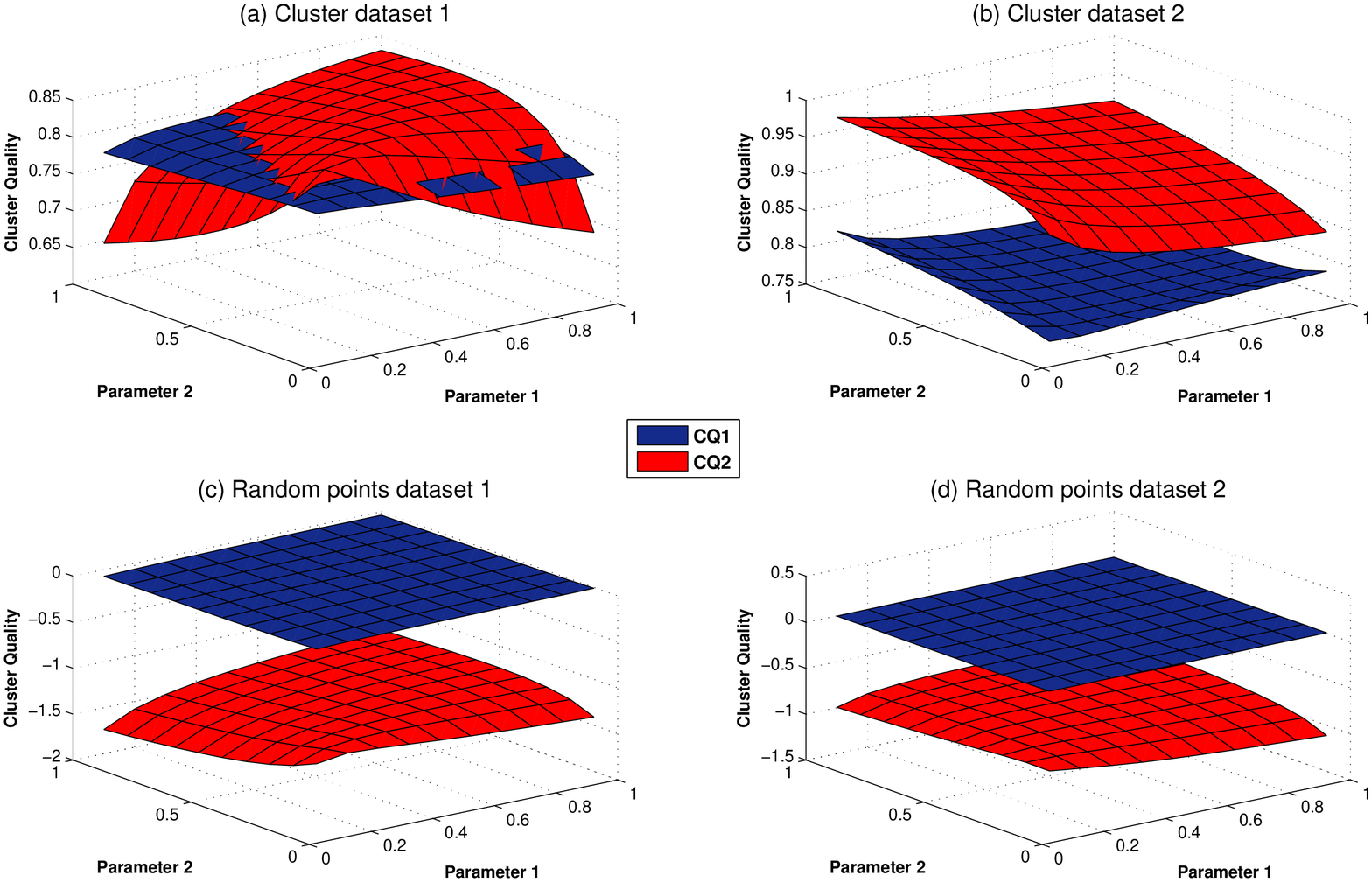}
 % condVScqProva.eps: 0x0 pixel, 300dpi, 0.00x0.00 cm, bb=0 0 1149 610
 \caption{With this picture we provide an example to justify our choice about the function used for evaluating the quality of a cluster. In particular, we plot the profile of the cluster quality computed using $\text{CQ}_1$ (in blue) or by considering $\text{CQ}_2$ (in red) when varying the PCs (2 parameters). In Fig. (a) and (b) we plot the profiles associated to two well-defined clusters in the two datasets in Fig. \ref{fig:ds_CQ}, while in Fig. (c) and (d) we plot the profiles of the cluster quality evaluated on some random points in the same datasets. $\text{CQ}_2$ shows a better discriminative power as it generally assumes a bigger range of values w.r.t. $\text{CQ}_1$, hence allowing for a better separation of relevant PCs. Furthermore, if we consider Fig. (a) and (c) or (b) and (d), it is possible to observe a more marked difference of the values associated to different clusters. In particular, if we use $\text{CQ}_2$ the difference of the values associated to good and bad clusters result to be higher 
w.r.t. the values obtained with $\text{CQ}_1$.}
 \label{fig:condvscq}
\end{figure}

From our preliminary experiments, we observed that the use of $\text{CQ}_2$ rather than $\text{CQ}_1$ characterizes much better the quality of the clusters in our multi-parameter setting.
To explain this fact with greater detail, let us consider an example where the two aforementioned functions are used for evaluating the cluster quality in two different datasets of $\mathbb{R}^2$ vectors depicted in Fig. \ref{fig:ds_CQ}.
We decided to consider two different datasets because the evaluation of the conductance is strictly correlated not only to the cluster itself, but also to the whole dataset to which it belongs.
In both datasets, we select two different subsets of vertices of the respective graph representations: the first one is associated with a well-defined cluster, while the second one is randomly determined, which accordingly induces a low quality cluster.
In Fig.~\ref{fig:condvscq} (a) and (b) we plotted the values assumed by $\text{CQ}_1$ and $\text{CQ}_2$ on the well-defined clusters, which are evaluated as a function of the PCs (in this case uniformly sampled in the parameters space $[0, 1]^2$).
Instead, in Fig.~\ref{fig:condvscq} (c) and (d) we performed the same calculations for the randomly determined clusters.
As it is possible to observe, the values assumed by $\text{CQ}_2$ fall within similar ranges in the two datasets, allowing to use comparable threshold values (i.e., $\tau_{\text{CQ}}$) for evaluating good clusters in different datasets.
By using the $\text{CQ}_2$ rather than $\text{CQ}_1$, we are also able to better discriminate those PCs that better characterize the clusters --  for the first dataset, these are individuated along the bisecting line, while for the second one PCs close to the $\{0, 1\}$ setting are preferable.
In fact $\text{CQ}_2$, in correspondence of such PCs, assumes values that better evaluate the quality of clusters: random clusters are always highly penalized while well-formed clusters are better magnified.

To conclude, since $\text{CQ}_2$ is normalized according to the conductance of the graph, we consider Eq. \ref{eq:CQ2} as an absolute quality measure that can be used for comparing clusters generated by different agents using different PCs. The soundness of such an assumption will be demonstrated by the experiments.

In the following of this paper we will always use $\text{CQ}_2$ as the function used for evaluating the quality of a cluster and, for the sake of notation, we will refer at it as ``CQ''. 

\subsection{Energy Update}
\label{sec:energyupdate}
Setting a proper value for the (maximum) length of a RW is another important issue to be considered, since it is strictly related to the typical size of the returned clusters/subgraphs.
A quantity called energy $e_i$ determines how many steps an agent $a_i$ is able to perform during a RW. The energy is initialized to a value $e_{\text{init}}$ and it is successively modified at each step of the RW.
As an agent visits the graph, it builds a subgraph $g_{hj}$ adding the new vertices that are being visited, increasing its size and modifying accordingly its current conductance.
In particular, when a vertex $v_l$ is inserted in $g_{hj}$, the conductance of the subgraph increases if $v_l$ is distant (i.e., very different in our setting) from the other vertices of $g_{hj}$; otherwise, the conductance will decrease.
Note that since the graph is complete, inserting a vertex $v_l$ to a subgraph $g_{hj}$ includes the insertion of all edges connecting $v_l$ to all vertices in $g_{hj}$.
Hence, the variation of the conductance during a RW can be used for discerning whether an agent is walking in the ``right'' or in the ``wrong'' direction, i.e., if the agent is visiting or not a compact area of $\mathcal{G}$.
For this reason, we modify the energy $e_i$ according to the variations of the conductance of $g_{hj}$ at each step of the RW: if the conductance is decreasing the energy increases, otherwise the energy is reduced.
If the conductance remains constant, it means that the agent is moving on vertices that have been already visited.
This happens when a suitably dense region has been completely visited and the agent is stuck moving on the same vertices over and over.
For this reason, we added also a constant energy decrement in order to consider loops that occur if the agent is not visiting new vertices for a prolonged period.

The expression describing the energy update reads as,
\begin{equation}
\label{eq:energy_update}
e_i^{(new)} = e_i^{(old)} + f(\Delta\phi(g_{hj})) - \tau_{\text{energy}},
\end{equation}
where $f(\Delta\phi(g_{hj}))$ is a function that depends on the variation of the conductance of $g_{hj}$, and $\tau_{\text{energy}}$ is the user-defined quantity that controls the constant decrement of the energy.
For evaluating the function $f(\cdot)$, we take into account how the conductance of the subgraph $g_{hj}$ varies each time a vertex is visited in the RW and, possibly, added to the subgraph. If the agent is correctly visiting the vertices of a proper cluster, we expect the conductance to decrease. This decrement, however, in most of the cases is neither regular nor monotone.
For this reason, we decided to consider an average computed on $r$ values estimated in $r$ steps of the RW.

The energy function $f\big[\Delta^r\phi(g_{hj})\big]$ that computes the moving average variation of the conductance on $r$ steps of the RW is defined as follows:
\begin{equation}
f\big[\Delta^r\phi\big(g_{hj}(t)\big)\big] = \frac{1}{r} \sum_{q = 0}^{r-1} \phi \big(g_{hj}(t-q) \big) - \phi\big(g_{hj}(t-q-1)\big),
\end{equation}
where $g_{hj}(t)$ indicates the subgraph $g_{hj}$ at the $t$-th time step of the RW.

If the value of $r$ is sufficiently low, $f\big[\Delta^r\phi(g_{hj})\big]$ quickly assumes a negative value when the agent exits from a cluster and then it is readily stopped. On the other hand, choosing a value too low for $r$, makes the system very sensitive to small variations of the conductance, which often occur when the agent is moving within the same cluster.
In our experiments (Sec. \ref{sec:exps}) we set $r=3$, a value that allows to detect sufficiently fast when an agent leaves a cluster, filtering at the same time non-relevant changes of the conductance.

\subsection{Selection of New PCs}
\label{sec:new_metrics}
A RW is terminated when the energy $e_i$ reaches a value lower or equal to zero. The subset of vertices that have been visited forms the resulting cluster, $c_{hj}$, whose quality is evaluated according to Eq. \ref{eq:CQ2}.
If $\mathrm{CQ}(c_{hj})$ is greater or equal than $\tau_{\text{CQ}}$, the cluster is added to the collection of good clusters discovered by the agents, along with the PC $m_j^{(i)}$ used by the agent for discovering such a cluster.
Since it is likely that a dataset contains more than one cluster of elements which are similar w.r.t. the same PC, it is reasonable to assume that if a PC $m_j^{(i)}$ has lead to the identification of a good cluster, it can be further exploited to discover additional good clusters within the same dataset.
Then, when a cluster is accepted we restore the initial quantity of energy of the agent, i.e. we set $e_i = e_{\text{init}}$ and we start a new RW on the same weighted graph, enabling the agent to explore a new unseen region of the graph.
For that reason, we set to zero the weights of the matrix $\mathbf{A}_j$ associated to the vertices which have been already visited by $a_i$ in the previous RW using $m_j^{(i)}$.
In this way both $\boldsymbol\pi_j$ and $\mathbf{M}_j$ are modified: changing $\boldsymbol\pi_j$ has the effect that the next RW starts from another dense region of the graph, while the modification of $\mathbf{M}_j$ prevents the agent from reaching vertices which have already been visited in the past.

Otherwise if $\mathrm{CQ}(c_{hj})$ is not high enough, $c_{hj}$ is rejected and the agent selects a new PC, say $m_{\text{new}}$; the energy $e_i$ is reset to the default starting value, $e_{\text{init}}$.
This implies the recalculation of $\mathbf{A}_{\text{new}}$, $\mathbf{M}_{\text{new}}$, and hence of $\boldsymbol\pi_{\text{new}}$, inducing a completely new RW characterized by a possibly different behavior.
The new PC $m_{\text{new}}$ is selected by considering a uniform distribution over $\mathcal{M}$.
In Sec. \ref{sec:exploration_exploitation} we describe a variant of LD-ABCD that implements a different PC selection strategy, which is more suitable for scenarios where the core dissimilarity measure is characterized by many parameters.

\subsection{Aggregation of Clusters/PCs}
\label{sec:cluster_aggr}
As long as the execution of LD-ABCD proceeds, an agent might find very similar (or even equal) clusters using different PCs, in the sense that they may overlap significantly. If an agent identify the same cluster $c_{h}$ using different PCs, $m_a$ and $m_b$, we say that such PCs are \textit{equivalent} w.r.t. $c_{h}$, in the sense that $c_{h}$ contains patterns that are characterized similarly by considering either $m_a$ or $m_b$.
This is an important qualitative information that describes the cluster in terms of the parameters of the dissimilarity measure used for discriminating the elements of the cluster from the rest of the dataset.
Additionally, showing that the same cluster can be obtained using different PCs underlines their relation within the dataset, allowing further analysis and semantic interpretations of the data at hand.

In order to group similar clusters, we merge into a single meta-cluster all such clusters whose intersection, in terms of contained patterns, is sufficiently high. It is therefore necessary  to define a dissimilarity measure among clusters: in order to do that, we represented each cluster $c_{hj}$ with a Boolean vector, $\mathbf{c}_{hj} \in \{0,1\}^n$, where each entry of the vector represents an index to an element in $\mathcal{S}$, in particular the $l$-th entry $\mathbf{c}_{hj}(l) = 1$ if the $l$-th pattern of $\mathcal{S}$ is contained in $c_{hj}$, while $\mathbf{c}_{hj}(l) = 0$ otherwise. At this point, the dissimilarity among clusters is computed though the the Hamming distance $d_H(\cdot,\cdot)$ that evaluates the distance among the two Boolean vectors that represent the clusters. Two clusters $c_1$ and $c_2$ are considered similar if their Hamming distance $d_H(c_1, c_2)$ is less or equal to $\theta$. The parameter $\theta\geq0$ is set proportional to $|\mathcal{S}|$ and it can be interpreted as the maximum fraction of patterns on which two clusters can disagree in order to be considered similar.

With $\hat{c}_{xi}$ we call the $x$-th meta-cluster associated to the agent $a_i$, that represents a set of clusters $\mathcal{C}_{xi}$ sufficiently similar to each other w.r.t. the Hamming distance. The meta-cluster $\hat{c}_{xi}$ is composed of a Boolean vector $\mu_{xi}$, defined as the rounded mean of all the Boolean representations of the clusters in $\mathcal{C}_{xi}$ and a list $\mathcal{L}_{xi}$ that contains all the PCs used for discovering the clusters in $\mathcal{C}_{xi}$. Each PC in $\mathcal{L}_{xi}$ is associated to a CQ value, which is used to perform a ``ranking'' of the PCs used for discovering the clusters; PCs associated to a meta-cluster are ordered in non-ascending order of CQ value.
In our experiments we have discovered that the PCs with higher CQ are the ones which better describe the original clusters in the dataset (see Sec. \ref{sec:relev_pc} and \ref{sec:eq_pcs}).

Every time a cluster $c_{hj}$ is discovered by an agent $a_i$ using a metric $m_j^{(i)}$, it is compared with all the mean Boolean vectors of the $K(t)$ meta-clusters existing at the time $t$, and it is assigned to the most similar meta-cluster, let say $\hat{c}_{xi}$. Then $c_{hj}$ is added to the set $\mathcal{C}_{xi}$ and $\mu_{xi}$ is recomputed on such set. Finally, the PC $m_j^{(i)}$ is added to $\mathcal{L}_{xi}$.

If no meta-clusters have still been generated, or if the dissimilarity value to the most similar meta-cluster is above a given threshold $\theta$, a new meta-cluster $\hat{c}_{yi}$ is instantiated starting from $c_{hj}$: in this case $\mu_{yi}$ is initialized with $c_{hj}$ and the metric $m_j^{(i)}$ used for discovering $c_{hj}$ is inserted in the list $\mathcal{L}_{yi}$, which initially will be empty.

With $\hat{\mathcal{C}}_i$ we refer to the collection of all the meta-clusters generated by $a_i$ which represents the set of all similar clusters that have been generated using different PCs.

When all the agents terminate their procedure of cluster discovery (see the following section), similar meta-clusters generated by different agents are in turn merged together by a centralized unit into a global meta-cluster. In fact, there are no guarantees that different agents do not generate the same meta-cluster. In order to aggregate 2 meta-clusters $\hat{c}_{a1}$ and $\hat{c}_{b2}$ generated by the agents $a_1$ and $a_2$, we check if the hamming distance between $\mu_{a1}$ and $\mu_{b2}$ is below the threshold $\theta$;  in that case  the clusters are merged in a new meta-cluster $\hat{c}_{\text{new}}$, where $\mathcal{C}_{\text{new}} =  \mathcal{C}_{a1} \cup \mathcal{C}_{b2}$, $\mathcal{L}_{\text{new}} =  \mathcal{L}_{a1} \cup \mathcal{L}_{b2}$ and $\mu_{\text{new}}$ is computed as the rounded mean element in $\mathcal{C}_{\text{new}}$.

\subsection{Convergence Criterion}
\label{sec:mamld_convergence}
To determine the convergence criterion of LD-ABCD we decided to analyze how the meta-clusters evolve, rather than considering the single clusters returned by the agents. In fact, due to the random nature of the walk, a single cluster returned by an agent may differ by very few elements from the already existing ones, making it hard to decide if it is an effectively new cluster.

The agents terminate the search when for a given time period, defined by the integer-valued threshold $\tau_{\text{stop}}$, no new meta-clusters are generated and the average cluster quality of the existing meta-clusters does not increase. The cluster quality of a meta-cluster is evaluated as the average of the cluster qualities of the single clusters associated to it.
In particular, if an agent returns consecutively for $\tau_{\text{stop}}$ times a cluster which is associated to an already existing meta-cluster and it does not improve its average CQ or if the cluster is rejected because its CQ is too low, the agent stops.
In fact, if an agent of the system has already visited the dataset with a sufficiently high number of PCs, it becomes less likely that new informative clusters are going to be discovered.
When all the agents reach their convergence criterion, the whole system stops and the results found by each agent are aggregated as described in the previous section.
The parameter $\tau_{\text{stop}}$ can be set by the user; it directly affects the execution time of the algorithm and accordingly the precision of the results.

\begin{figure}[ht!]
 \centering
 \includegraphics[bb=0 0 778 430,scale=0.5,keepaspectratio=true]{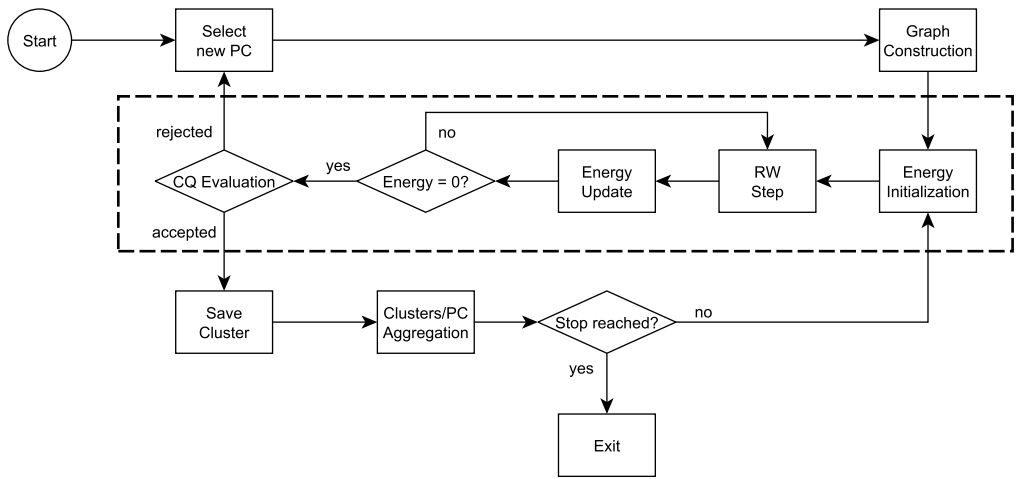}
 % schema2.eps: 0x0 pixel, 300dpi, 0.00x0.00 cm, bb=0 0 778 430
 \caption{Detailed flow-chart of the single agent behavior.. The diagram part enclosed in dashed line is the expansion of the ``Perform RW and Evaluate Cluster'' block of Fig. \ref{fig:schema1} and it shows more in detail the sequence of operations that an agent $a_i$ performs during the random walk.. The energy $e_i$ is initialized to a default value $e_{\text{init}}$. At each step a new node is considered and, if it is visited for the first time by the agent, it is inserted in the cluster. The energy of the agent is modified accordingly to the variation of the conductance of the cluster found so far. When the energy is depleted, the cluster is evaluated and if its cluster quality is sufficiently high it is accepted and saved, otherwise it is discarded and a new PC is selected.}
 \label{fig:schema2}
\end{figure}

Now that all the functionalities of the system have been explained, we present in Fig. \ref{fig:schema2} a more detailed overall-schema of a single agent behavior over its lifetime.

\subsection{Analysis of Computational Complexity}
\label{sec:compcomplexity}

In this section we study the time and space complexity of LD-ABCD. For what concerns the space occupancy, the upper-bound consists in storing the weighted adjacency matrix, $\mathbf{A}$, which each agent must use in order to represent the graph. The space required to store the matrix is $O(n^2)$, where $n=|\mathcal{S}|$.

On the other hand, the time complexity strictly depends on the number of iterations performed by each agent during the random walk.
The length of a typical RW is related to the energy $e$ of the agent and on how this quantity is modified (which is affected by the experimental setting of the algorithm and by the intrinsic random nature of the RWs). The energy variation depends also on the nature of the dataset at hand, which makes a precise analysis difficult to perform.
In order to give an estimation of the computational time complexity, we assume here that an agent performs in average $T$ different steps during a typical RW. 

The time complexity can be estimated as the composition of several costs.
The operations performed by an agent can be divided in the following categories, which scale with the input data size in different ways: 

\begin{itemize}

\item{the \textit{PC initialization} step, which includes the generation of the PC given the selected policy and the evaluation of the adjacency and transition matrices and the computation of the graph conductance bounds. Sampling a random PC has a cost that scales linearly with the number of parameters of the dissimilarity measure, and so it can be generally considered negligible w.r.t. the costs depending on dataset size. Building the adjacency and transition matrices has a cost of $O(n^2 \cdot \delta)$, where $\delta$ is the cost of the dissimilarity measure. Evaluating the bounds of the graph conductance, used for evaluating CQ (see Eq. \ref{eq:CQ2}), has the same cost of computing the second eigenvalue of the adjacency matrix, which in our study it has been approximated with the power method described in Appendix~\ref{sec:graph_conductance}. The power method complexity scales as $O((n + n^2) \cdot \frac{1}{\epsilon} \cdot \log \frac{n}{\epsilon})$, where $\epsilon$ is the user-defined parameter defining 
the precision on the approximation. We refer then to the time required for initialization step with $t_{\text{init}} = O(n^2 \cdot (\delta + \log(n)))$;}

\item{the \textit{random walk} step, which consists in selecting a new node and updating the energy $e$ of an agent, according to the variation of the conductance of the subgraph visited so far. While the energy updating procedure can be performed in a constant time, selecting the next node in the RW is an operation which involves analyzing all the elements of the row of the adjacency matrix relative to the current node, which scales as $O(n)$. We then define the cost $t_{\text{step}} = O(n)$;}

\item{the \textit{cluster quality evaluation} step consists in the evaluation of the cluster conductance, which is an operation that costs $O(n^2)$, since all the edges of the (complete) graph must be considered -- see Appendix \ref{sec:graph_conductance}. The estimated time required for performing this step is given by $t_{\text{eval}} = O(n^2)$.}

\item{the \textit{cluster aggregation} step that consists in updating the set of existing meta-clusters with the cluster that has been accepted by the agent. This operation consists in comparing the cluster with all the $K(t)$ meta-clusters which have been generated so far at the time $t$, using the hamming distance. The hamming distance is linear in the number of the elements, which is the size of the dataset $n$, since each cluster is represented in the vectorial form described in Sec. \ref{sec:cluster_aggr}. Note that the aggregation procedure occurs only when a cluster is accepted, i.e. when its quality is sufficiently high, so this cost sometimes is equal to zero. We can then define $t_{\text{aggr}} = O(K(t) \cdot n)$.}

\end{itemize}

To summarize, the total time $t_{\text{tot}}$ required by an agent to evaluate a PC $m_j$ can be expressed as:
\begin{align*}
 t_{\text{tot}} =& t_{\text{init}} + T \cdot t_{\text{step}} + t_{\text{eval}} + t_{\text{aggr}} \\
 =& O(n^2 \cdot (\delta + \log(n))) + T \cdot O(n) + O(n^2) + O(K(t) \cdot n)
\end{align*}

Since the procedure must be repeated each time a new PC is considered, the total time required for executing the whole LD-ABCD system is $M \cdot t_{\text{tot}}$, where $M$ is the number of PC evaluated (we remind that the number of agents is fixed in our algorithmic setting).

%%%%%%%%
\section{LD-ABCD with Exploration--Exploitation Agents}
\label{sec:exploration_exploitation}

Since the PC space can be extremely large even for a modest number of parameters of the dissimilarity measure, the technique used for searching PCs described in Sec. \ref{sec:new_metrics} -- uniform sampling -- could easily become ineffective. In this section, we propose an alternative approach for exploring the PC space.
The search method is inspired to the well-known Metropolis-Hastings algorithm \cite{metropolis}, often employed in statistical physics.
In this variation, the agents operate according to two different policies (strategies, behaviors), which we named \textit{exploration} and \textit{exploitation}.
An agent that operates according to the exploration strategy is called ``explorer''.
The exploration strategy coincides with the uniform search described in Sec. \ref{sec:new_metrics} and it is meant to perform an exploratory wide-range search in the PC space. An explorer randomly evaluates several different PCs. Every time a RPC is identified by an explorer, it is stored to a shared data structure to allow successive tentative improvements via the exploitation.
Accordingly, an agent that implements the exploitation strategy, instead, is called ``exploiter''. The objective of the exploiters consists in trying to improve the RPCs found so far by the explorers.
An exploiter randomly selects one of the available RPCs, say $m_j^{(i)}$, along with its corresponding cluster $c_{hj}$, and initiates a search in the PC space nearby $m_j^{(i)}$, given a suitable PCs similarity measure $d_{\text{PC}} (\cdot, \cdot)$. 
This search strategy is meant to discover other PCs that yield a higher CQ (\ref{eq:CQ2}) on the same cluster $c_{hj}$.
In fact, since it is reasonable to assume that agents with similar PCs are likely to perform similar RWs (and hence accept/reject similar clusters), we keep fixed the cluster structure (i.e., the patterns that it contains) and we just recompute its CQ using the new PCs.
The fact that we recompute the CQ of the cluster without issuing a new RW results in a significant improvement in terms of computational resources.
The implementation of the similarity measure between PCs depends on the nature of the parameters (e.g., Hamming distance for binary configurations, Euclidean distance for real-valued parameters, etc.).
If an exploiter is able to select a new PC $\bar{m}_j^{(i)}$ that yields a better CQ than $m_j^{(i)}$, this latter is deleted (along with the related cluster $c_{hj}$) and it is replaced by $\bar{m}_j^{(i)}$ and the associated cluster by $\bar{c}_{hj}$.

\begin{figure}[ht!]
 \centering
 \includegraphics[bb=0 0 456 607,scale=0.5,keepaspectratio=true]{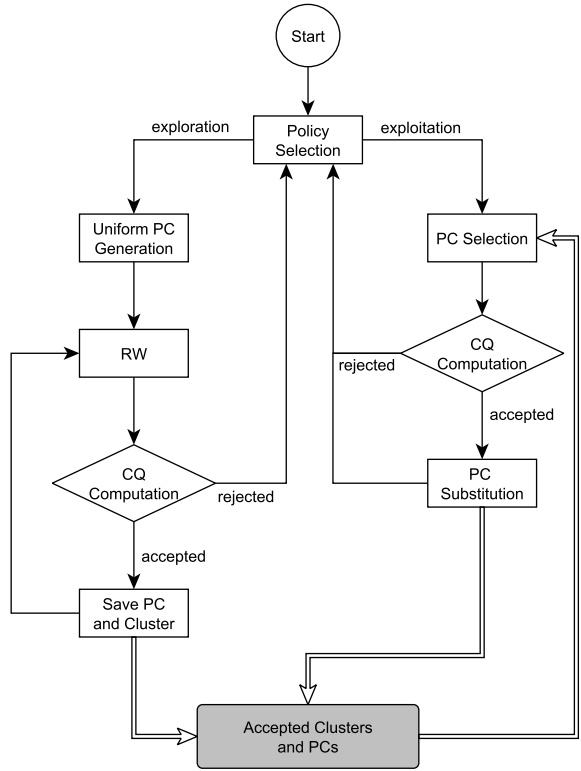}
 % exploitschema.eps: 0x0 pixel, 300dpi, 0.00x0.00 cm, bb=0 0 456 607
  \caption{Diagram of the Exploration--Exploitation procedure for selecting the new PC. The thick arrows represent the read/write operations performed by the agents on the shared data structure, highlighted in gray, containing the discovered clusters and PCs.}
\label{fig:exploitschema}
\end{figure}

Every agent can exclusively assume the role of the explorer or the exploiter (Fig. \ref{fig:exploitschema}), modifying hence its search strategy accordingly.
Before starting a new RW, an agent checks the current ratio of explorers and exploiters operating in the system.
If the ratio is above a user-defined threshold $0 < \tau_{\text{EXPL}} \leq 1$, and at least one RPC has been already discovered by an explorer, the agent adopts the exploitation policy, otherwise it behaves as an explorer.
The factor $\tau_{\text{EXPL}}$ controls the balance between the diversity and the accuracy of the returned RPCs and can be tuned according to the available computational resources and the particular problem at hand.
The exploration--exploitation version of LD-ABCD herein discussed is designed to be able to perform a more targeted search on large PC spaces. This results, in general, in a faster convergence of the whole algorithm, with a faster discovery of the high-quality clusters and related PCs present in the data (we will provide experimental evidence of this claim later in Sec. \ref{sec:exploration_exploitation_tets}).
Finally, the herein presented exploration-exploitation variant is characterized by the same computational costs described in Sec. \ref{sec:compcomplexity}, as the operations for the explorers and exploiters are asymptotically the same.

%%%%%%%%
\section{Experiments}
\label{sec:exps}

In this section we discuss the experiments performed to asses the performances of (both variants of) LD-ABCD.
First, in Sec. \ref{sec:rw_purity} we discuss the tests performed to evaluate the quality of the clusters found by LD-ABCD on some well-known benchmarking datasets. We offer a comparison w.r.t. state-of-the-art graph-based and RW-based algorithms over a particular setting of clustering, where patterns are labeled with ground-true class labels for performance evaluation.
Then in Sec. \ref{sec:relev_pc} we present some experiments which underline the capability of our system to discover relevant information in noisy datasets. Notably, the identification of relevant clusters, together with the PCs used for discovering such clusters, provide a semantic characterization and a high-level description of the data.
In Sec. \ref{sec:eq_pcs} we demonstrate the capability of LD-ABCD to discover multiple PCs which characterize individual clusters, defining then a relation among the features considered by each PCs in the data contained in the cluster.
Those first three experiments are performed with the LD-ABCD version discussed in Sec. \ref{sec:algodesc}.
Finally, in Sec. \ref{sec:exploration_exploitation_tets} we discuss the results obtained by using the exploration--exploitation technique described in Sec. \ref{sec:exploration_exploitation} for improving the selection (discovery) of the RPCs.

As stressed throughout the paper, our approach is dissimilarity-based. Therefore LD-ABCD is able to process virtually any input data type (e.g., graphs, sequences and so on).
However, for the sake of simplicity and for an easier interpretation of the results, we decided to test only datasets of real-valued vectors (features); extensions to other settings are straightforward.
The adopted dissimilarity measure is the weighted Euclidean distance; each $m_j^{(i)}$ is a vector in $[0, 1]^D$, where $D$ is the dimensionality of the data at hand. We do not use a Mahalanobis-like distance (e.g., by using the full weight matrix), since the former distance allows a more direct interpretation of the results in terms of feature selection (and it is characterized by much less parameters).

Our algorithm depends on a number of parameters and thresholds, which are $\tau_{\text{CQ}}$, $\tau_{\text{stop}}$, $\beta$ (used in the definition of $\tau_{\text{exp}}$), $\theta$ and $\tau_{\text{energy}}$. In our experiments we used different configurations of those parameters, that have been set empirically accordingly to the dataset and to the problem at hand. However, in several cases we kept those parameters unaltered, since modifying their values does not lead to any remarkable changes in the results, making their choice not very critical.

\subsection{Evaluating the Purity of the RWs}
\label{sec:rw_purity}

We have processed four different real-world datasets from the UCI Machine Learning Repository \cite{Bache+Lichman_2013}, which are \emph{Wine}, \emph{Breast Cancer}, \emph{Iris}, and \emph{E-Coli}. We decided to use the aforementioned datasets since they are very well-known, easy to obtain and for some of them it was possible to provide a comparison with the results obtained by other algorithms which perform clustering using a RW \cite{5693955}.
All datasets contain labeled patterns organized in different classes.
As we described in Sec. \ref{sec:clusterquality}, the LD-ABCD algorithm uses the $\mathrm{CQ}$ (\ref{eq:CQ2}) -- a criterion based only on the evaluation of the conductance -- for accepting or rejecting the clusters identified during the RWs.
In the following experiments, we demonstrate the reliability of our (unsupervised) cluster acceptance criterion using the supervised information of the class labels. In this test, the PCs are defined as real-valued numbers.

As mentioned before, we provide a comparison with the MARW algorithm \cite{5693955} and two other algorithms therein considered, which are \emph{Nibble} \cite{DBLP:journals/corr/abs-0809-3232} and \emph{Apr.PageRank} \cite{andersen2006local} (in the following denoted as N and APR), relatively to the first two dataset treated (Wine and Breast Cancer).
MARW is an agent-based and RW-based clustering algorithm. Agents perform the RW on the same graph together, with the constraint of having a (geodesic) distance of at most $l$ from each other. This corresponds to decreasing the chance that the multi-agent RW ``mistakenly'' merges two different clusters (low transition probabilities are easily zeroed).
To make results comparable, we adopted the same performance measure described in \cite{5693955} for evaluating the purity of a cluster. The purity is the percentage of vertices visited during the RW that has the same class label of the starting vertex.
Let $v_s$ be the starting vertex, $l(v)$ the true label value of $v$, and $c_h$ the accepted cluster made of vertices visited during a RW.
The cluster purity (CP) is defined as:
\begin{equation}
\label{eq:cp}
 r = \frac{|\{ v | l(v) = l(v_s) \}|}{|c_h|}.
\end{equation}

In LD-ABCD, the starting node, $v_s$, is selected from the SD, $\boldsymbol\pi$ (see Sec. \ref{sec:random_walk_cluster}). Hence, $v_s$ is selected from a central part of the graph, making its class label a reliable estimation of the class of the cluster to which $v_s$ effectively belongs.

For each processed dataset, we identify $K$ meta-clusters and their associated collection of equivalent PCs (see Sec. \ref{sec:cluster_aggr}). From each meta-cluster, we chose the PC that has generated the cluster with the highest CQ and then we check its CP (\ref{eq:cp}). We use the average value of those $K$ CPs as the performance index on the whole dataset (we report the standard deviations).
The results obtained by our system are reported in Tab. \ref{tab:uci}, along with the results found by the other algorithms for what concerns the first two datasets.
\begin{table}[thp!]\tiny
\begin{center}
\caption{CP results on the considered UCI datasets. In parentheses we show the number of distinct vertices required to stop the RWs.}
\label{tab:uci}
\begin{tabular}{|c|c|c|c||c|c|c|c|c|}
\hline
\multicolumn{4}{|c||}{\textbf{Datasets}} & \multicolumn{5}{c|}{\textbf{Algorithms}} \\
\hline
 & \textbf{Patterns} & \textbf{Dimensions} & \textbf{Classes} & \textbf{N} & \textbf{APR} & \textbf{MARW $a=3$} & \textbf{MARW} $a=4$ & \textbf{LD-ABCD} \\
\hline
Wine (40) & 178 & 13 & 3 & 82.31 & 86.80 & 88.02 & 91.76 & 100.0$\pm0.000$ \\
\hline
Breast Cancer (160) & 683 & 9 & 2 & 93.26 & 94.66 & 94.37 & 96.02 & 100.0$\pm0.000$ \\
\hline
Iris (40) & 150 & 4 & 3 & - & - & - & - & 76.00$\pm0.120$ \\
\hline
E-Coli (50) & 336 & 8 & 8 & - & - & - & - & 91.00$\pm0.231$ \\
\hline
\end{tabular}
\end{center}
\end{table}

\begin{figure}[ht!]
 \centering
 \includegraphics[bb=0 0 713 607,keepaspectratio,scale=0.55]{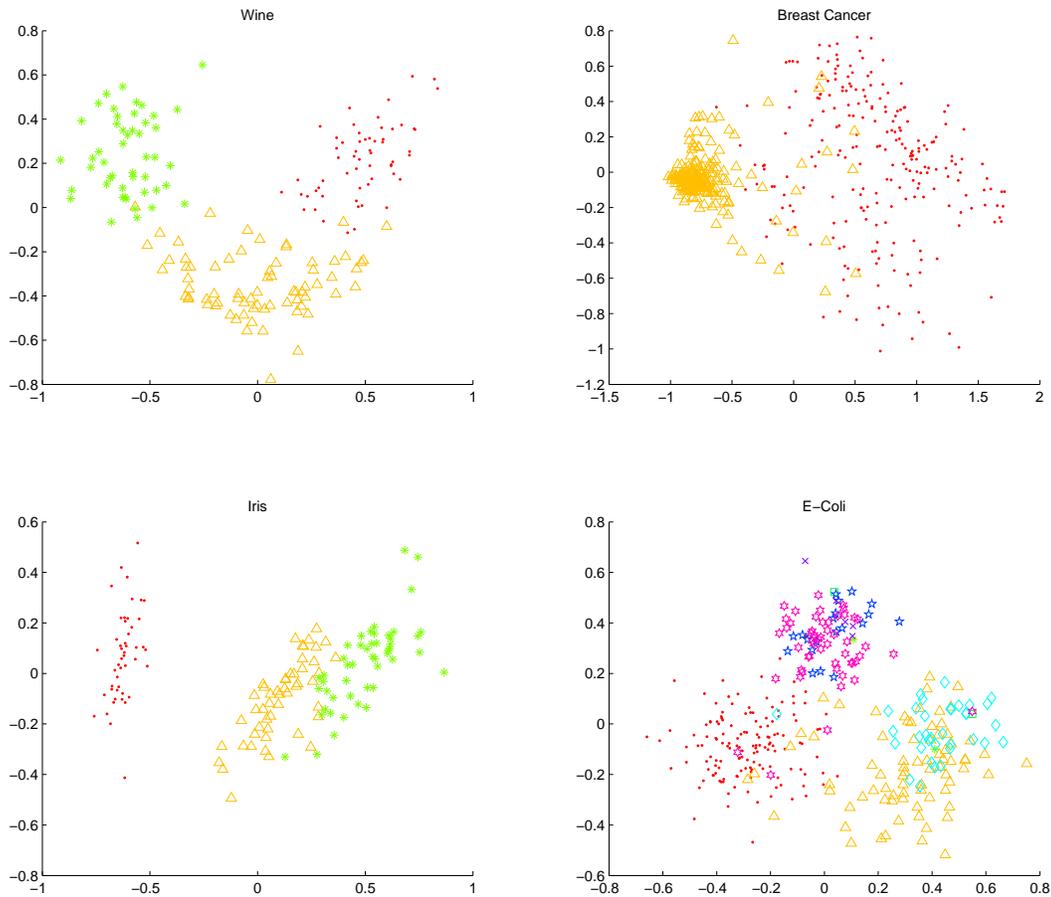}
  \caption{First two components of the PCA of the considered UCI datasets. We use different colors and shapes to distinguish among the different classes in the dataset (color version online).}
 \label{fig:PCADatasets}
\end{figure}

In addition to this numerical comparison, in the following we briefly discuss the behavior of LD-ABCD in each dataset, in order to provide a more complete overview of its functioning. Since there is no pre-processing on the considered data, we decided to show a principal component analysis (PCA), which we use only for facilitating the comprehension of the following discussion (see Fig. \ref{fig:PCADatasets}).

\paragraph{Wine}
In this dataset, LD-ABCD was able to identify three different meta-clusters that correctly cover the three classes of the dataset. Each meta-cluster contains only patterns belonging to a single class and thus the CP associated to the PC with the highest CQ is 1 in every meta-cluster. According to MARW \cite{5693955}, we stopped the RWs as soon as a given number $z$ of \emph{different} vertices are visited. The value of $z$ is selected proportional to the smallest class in the dataset at hand.

\paragraph{Breast Cancer}
This dataset contains two different classes of patterns which are characterized by a very different distribution, as we can see from the related PCA in Fig. \ref{fig:PCADatasets}. The elements of the first class are very similar and they occupy a compact portion of the space, while the others are spread on a less dense region. On this dataset our algorithm returned only one meta-cluster containing patterns belonging exclusively to the first class and thus the resulting average CP is 1.
If at a first sight the absence in the output of a meta-cluster representing the second class may look as a failure, this behavior is perfectly aligned with the design of LD-ABCD, which tries to identify only the most compact and separated clusters in the dataset.
From the point of view of clustering, in this dataset there is only one well-defined cluster (those in blue). In fact, every time an agent tries to evaluate a cluster over the red patterns, it systematically rejects those clusters because the related CQ is too low (they are highly conductive).

\paragraph{Iris}
For this dataset we have performed two different runs. In the first one, we have kept the threshold $\tau_{\text{CQ}}$ to the standard value ($0.9$) used in all other experiments, while in the second run we have lowered it to $0.5$, In this way we allowed the algorithm to return more clusters, since the ones with a lower CQ are accepted. In the first test, only one meta-cluster is returned that contains points from the most isolated region (see the PCA in Fig. \ref{fig:PCADatasets}). The CP obtained in this first run is equal to 1. 

In the second run, instead, three different meta-clusters are returned. The first one contains again elements of the most isolated class and its CP is equal to 1, while the others two meta-clusters represent the two remaining classes and their CP is lower.
In fact they are not well-separated and the agent during a random walk switch between elements belonging to these two different classes, decreasing the CP of the resulting clusters. Also the CQ of those two clusters is significantly low since the agent moves freely on a larger portion of the graph, returning then a subgraph characterized by a higher conductance. The CP associated to the PC with the highest CQ of those two clusters is respectively $0.6$ and $0.67$, making the total CP obtained on the dataset equal to $0.76$.

\paragraph{E-Coli}
Notwithstanding the dataset contains 8 different classes, the number of the resulting meta-clusters is 3 and they are mainly populated by patterns belonging to the largest classes of the dataset. In fact, the number of elements in the 5 remaining classes is remarkably lower, and they have been partially aggregated in the clusters representing the 3 principal classes. For this reason, the CP obtained on this dataset is not 1, even if it still maintains a good score: the best PC of the 3 clusters have the following CP: $0.96, 0.93$ and $0.84$, making the average CP of the whole dataset $0.91$.

\subsection{Discovering Relevant PCs}
\label{sec:relev_pc}

In this first test, we focus on the problem of finding the PCs that best highlight the local structure of the clusters characterizing the dataset.
In particular, we identify a collection of meta-cluster (see Sec. \ref{sec:cluster_aggr}) associated with the list $\mathcal{L}$ of the equivalent PCs that have been used for identifying the aggregated clusters. We order the PCs in $\mathcal{L}$ according to their CQ (see Sec. \ref{sec:clusterquality}).
Since by definition each cluster of the considered dataset is characterized by its own specific PC, we expect (i) to retrieve the correct PC and (ii) that the PC associated to the highest CQ is the one that better characterizes the cluster.

In order to demonstrate the capabilities of LD-ABCD, we have generated a synthetic dataset in $\mathbb{R}^4$, which contains 4 different clusters $c_1$, $c_2$, $c_3$, and $c_4$. The vectors forming each cluster are characterized by values drawn from a tensor product of a three-dimensional Gaussian distribution with spherical covariance matrix and a unidimensional uniform distribution -- the uniform distribution plays the role of the noise.
For each of the 4 clusters, we select a specific dimension to add the values that come from a uniform distribution.
Specifically, referring with $x[n]$ as the $n$-th component of the vectors of the dataset, we insert the values drawn from the uniform distribution in $x[1]$ relatively to the patterns of $c_1$, in $x[2]$ for the patterns of $c_2$, in $x[3]$ for the patterns of $c_3$, and finally in $x[4]$ for the patterns of $c_4$.

\begin{figure}[ht!]
 \centering
 \includegraphics[bb=0 0 359 272,scale=0.7,keepaspectratio=true]{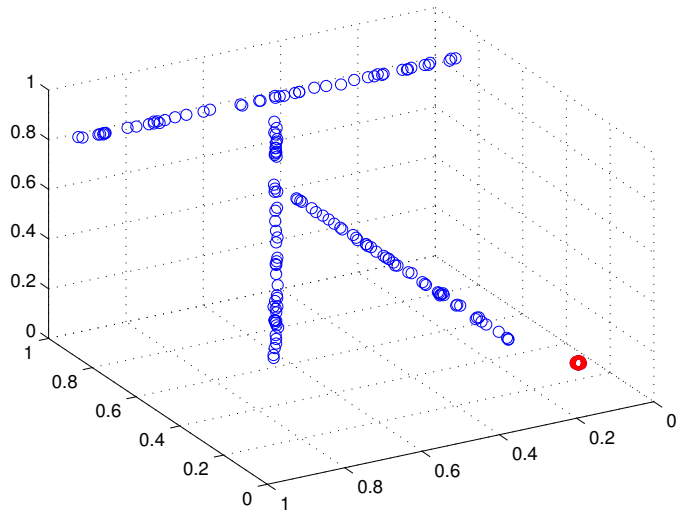}
 % 4d4c.eps: 0x0 pixel, 300dpi, 0.00x0.00 cm, bb=0 0 359 272
  \caption{Plot of the first three dimensions of the dataset characterized by four clusters in $[0, 1]^4$. Every cluster contains vectors with a component whose values are drawn from a uniform distribution, which plays the role of a noisy component. The blue clusters have that component in one of the displayed dimensions, while the 4-th dimension of the red cluster is the one containing the noise.}
\label{fig:4d4c}
\end{figure}

In Fig. \ref{fig:4d4c} we show the first three components of the considered patterns, omitting the 4-th component, $x[4]$. As it is possible to observe from the figure, although the clusters are characterized by a narrow variance on a specific dimension, they are clearly well-separated.
While the clusters $c_1$, $c_2$, and $c_3$ (plotted with blue dots) have the component containing the noise in one of the three displayed dimensions (respectively on $x[1]$, $x[2]$, and $x[3]$), $c_4$ (plotted with red dots) has all the components with values drawn from the Gaussian distribution in $\mathbb{R}^3$ and the component containing the noise is $x[4]$.
Note that the values of $x[4]$ for the blue clusters are drawn from a Gaussian distribution instead.
We execute LD-ABCD using Boolean PCs only ($m_j$ are Boolean vectors), until the stop criterion (described in Sec. \ref{sec:mamld_convergence}) is reached. As expected, LD-ABCD discovered four different meta-clusters $\hat{c}_{\text{i}}$, $\hat{c}_{\text{ii}}$, $\hat{c}_{\text{iii}}$, and $\hat{c}_{\text{iv}}$.
In Tab. \ref{tab:4d4c} we report the PC with higher CQ found for each meta-cluster and the relative CQ value.
\begin{table}[thp!]\scriptsize
\begin{center}
\caption{PCs found for the four meta-clusters and the associated CQ values.}
\label{tab:4d4c}
\begin{tabular}{|c|c|c|}
\hline
\textbf{Meta-cluster} & \textbf{PC} & \textbf{CQ} \\
\hline
$\hat{c}_{\text{i}}$ & $\{ x[1], x[2], x[3], x[4] \} = \{ 0, 1, 1, 1 \}$ & 0.9247 \\
\hline
$\hat{c}_{\text{ii}}$ & $\{ x[1], x[2], x[3], x[4] \} = \{ 1, 0, 1, 1 \}$ & 0.9442 \\
\hline
$\hat{c}_{\text{iii}}$ & $\{ x[1], x[2], x[3], x[4] \} = \{ 1, 1, 0, 1 \}$ & 0.9181 \\
\hline
$\hat{c}_{\text{iv}}$ & $\{ x[1], x[2], x[3], x[4] \} = \{ 1, 1, 1, 0 \}$ & 0.9475 \\
\hline
\end{tabular}
\end{center}
\end{table}

As it is possible to observe, the PCs that have been found showing the highest CQ values are those that assign $1$ in each cluster in correspondence of the components drawn from the Gaussian distribution (i.e., the signal), and $0$ to the component drawn from uniform distribution (i.e., the noise).
This demonstrates that LD-ABCD is able to discover the local structure of the relevant clusters in the dataset, identifying also the specific PC that allow such structures to emerge.

In our experiment we reported for each cluster the first PC in the list $\mathcal{L}$ of equivalent PCs, that is the one with the highest CQ value, and thus the one that better characterizes the cluster. Such PCs are reported in Tab. \ref{tab:4d4c}. Notice that for this test the threshold $\tau_{\text{CQ}}$ can be set to an arbitrarily low value, because we are considering only the first PC (in terms of CQ) in $\mathcal{L}$ and ignoring the others.

\subsection{Identification of Equivalent PCs}
\label{sec:eq_pcs}

In this section we evaluate the capability of LD-ABCD to discover the PCs which can equivalently characterize a portion of the data.
In Sec. \ref{sec:cluster_aggr} we have introduced the concept of equivalent PCs which are associated to each meta-cluster. Such PCs are collected in the structure $\mathcal{L}$ associated to each meta-cluster in $\hat{\mathcal{C}}$. Each PC in $\mathcal{L}$ is characterized by a specific CQ value: the higher the CQ, the better the PC characterizes the meta-cluster.
If a meta-cluster is associated with a set of PCs that are characterized by high and similar CQ values, we interpret them as equivalent, in the sense that they can be used interchangeably to suitably identify and characterize locally the cluster.
Furthermore, we can identify relations among the parameters w.r.t. the dataset at hand.

\begin{figure}[ht!]
 \centering
 \includegraphics[viewport=0 0 498 482,scale=0.4,keepaspectratio=true]{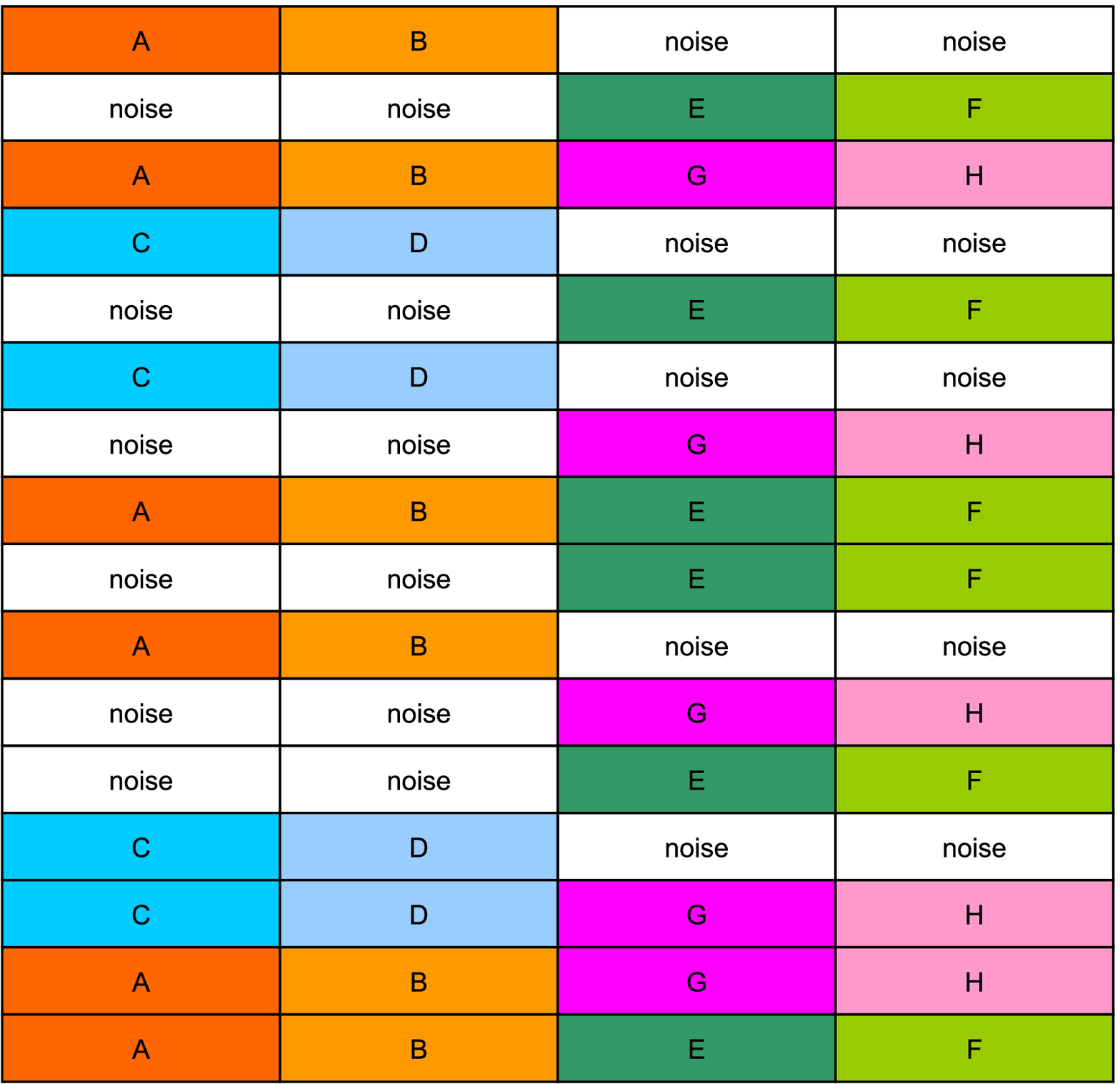}
 \caption{(Color version online) A snapshot of the dataset of vectors in $[0, 1]^4$, containing four different clusters. Vectors of a given cluster are characterized by two components (signal) drawn from 8 different Gaussian distributions $G_A,..,G_H$ (characterized in the figure by the use of different colors) and two components which are not relevant. For example, the vectors of the first cluster have values drawn from $G_A$ on the first component, values drawn from $G_B$ on the second component, and random values on the last two components.}
 \label{fig:eq_pcs}
\end{figure}

To show this process and make it easily understandable, we have used a synthetic dataset in $[0, 1]^4$, which contains four different clusters.
Each cluster contains data points which are very compact in two dimensions, while having uncorrelated values in the other two dimensions.
More precisely, the projection of the cluster on the hyperplane formed by the first two dimensions is normally distributed with narrow variance around the center. This means that the first cluster is defined by the vectors whose first two components are extracted from two Gaussian distributions, $G_A$ and $G_B$; the second cluster is formed by vectors whose third and fourth components are drawn from the distributions $G_E$ and $G_F$, and so on (see Fig. \ref{fig:eq_pcs} for an illustration).
On the remaining dimensions, the vectors contain values which are drawn from a mixture of different Gaussian distributions (each one belonging to a different cluster) or noise.
Since we wanted to keep the data in each cluster sufficiently isolated from the others, we drew the noise values by a random sampling considering a domain obtained by subtracting from $[0, 1]^4$ a suitable neighborhood of all the clusters.

In this sense, each cluster can be identified by PCs which assign high weights to any of the two signal components (or both), and a low weight to the others.
For example, if we consider Boolean PCs, the cluster which contains vectors of the type $[A,B,\sim,\sim]$, where $\sim$ denotes either a signal different from A and B or a noisy component, can be identified by the following equivalent PCs: $\{1,1,0,0\}$, $\{0,1,0,0\}$, and $\{1,0,0,0\}$.

In the herein presented experiment, we have generated a dataset of the form described above, which is exemplified in Fig. \ref{fig:eq_pcs}. Such a dataset contains 300 vectors in $[0, 1]^4$, whose components are real values extracted from eight different Gaussian distributions $G_A, G_B, ..., G_H$ or from a uniform distribution.
Each Gaussian distribution is paired with another one, in the sense that if a vector contains a value extracted from a distribution, it must also contain a value extracted from a second one. For example, the vectors which have the first component extracted from $G_A$ must have the second component extracted from $G_B$, while the remaining two components can contain any other value.
In this way we assure a correlation between pairs of Gaussian distributions for each specific cluster. This fact is illustrated in Tab. \ref{tab:eq_pcs1}.
\begin{table}[thp!]\scriptsize
\begin{center}
\caption{characterization of the clusters of the dataset in terms of vector components extracted from Gaussian distribution.}
\label{tab:eq_pcs1}
\begin{tabular}{|c|c|c|c|c|}
\hline
\textbf{Cluster} & \multicolumn{4}{c|}{\textbf{Vectors components}} \\
\hline
$c_1$ & $G_A$ & $G_B$ & $\sim$ & $\sim$ \\
\hline
$c_2$ & $G_C$ & $G_D$ & $\sim$ & $\sim$ \\
\hline
$c_3$ & $\sim$ & $\sim$ & $G_E$ & $G_F$ \\
\hline
$c_4$ & $\sim$ & $\sim$ & $G_G$ & $G_H$ \\
\hline
\end{tabular}
\end{center}
\end{table}

The mean of each Gaussian component is randomly generated, constrained to be separated by the others by a value greater or equal to $0.2$; we used a variance $\sigma = 0.005$. The subspace from which we draw the noisy values using the uniform distribution is defined by setting a radius $0.1$ for the neighborhoods of the clusters that we subtract from $[0, 1]^4$.

The results obtained by running the system with the cluster quality threshold $\tau_{\text{CQ}}$ equal to $0.8$ are reported in Tab. \ref{tab:eq_pcs2}.
As it is possible to observe, six different meta-clusters, $\hat{c}_{\text{i-vi}}$, have been found and four of them, $\hat{c}_{\text{i,ii,v,vi}}$, correspond, respectively, to the expected clusters $c_{2,1,3,4}$, while the two remaining meta-clusters, $\hat{c}_{\text{iii,iv}}$, correspond to high density areas that occurred randomly in the generation of the dataset.
As Tab. \ref{tab:eq_pcs2} shows, the meta-clusters $\hat{c}_{\text{i,ii,v,vi}}$ have associated PCs which select the relevant component of the vectors, according to the way the clusters have been generated.
\begin{table}[thp!]\scriptsize
\begin{center}
\caption{Meta-clusters found with the respective list of the first three PCs sorted in terms of CQ associated to each cluster. For each meta-cluster, we report also the ground truth cluster with which it intersects the most and the related cardinality of the intersection expressed in percentage.}
\label{tab:eq_pcs2}
\begin{tabular}{|c|c|c|c|}
\hline
\textbf{Meta-Cluster} & \textbf{Cluster with Highest Intersection} & \textbf{PCs} & \textbf{CQ} \\
\hline
\multirow{3}{*}{$\hat{c}_{\text{i}}$} & \multirow{3}{*}{$c_2 (100 \%)$} & $[1,0,0,0]$ & $0.895$ \\
\cline{3-4}
&  & $[1,1,0,0]$ & $0.876$ \\
\cline{3-4}
&  & $[0,1,0,0]$ & $0.859$ \\
\hline
\multirow{3}{*}{$\hat{c}_{\text{ii}}$} & \multirow{3}{*}{$c_1 (100 \%)$} & $[1,0,0,0]$ & $0.877$ \\
\cline{3-4}
&  & $[0,1,0,0]$ & $0.856$ \\
\cline{3-4}
&  & $[1,1,0,0]$ & $0.851$ \\
\hline
\multirow{3}{*}{$\hat{c}_{\text{iii}}$} & \multirow{3}{*}{$c_2 (26 \%)$} & $[1,1,1,1]$ & $0.902$ \\
\cline{3-4}
&  & $[1,0,1,0]$ & $0.890$ \\
\cline{3-4}
&  & $[0,1,1,0]$ & $0.817$ \\
\hline
\multirow{3}{*}{$\hat{c}_{\text{iv}}$} & \multirow{3}{*}{$c_4 (22 \%)$}  & $[1,1,1,1]$ & $0.882$ \\
\cline{3-4}
&  & $[1,1,1,0]$ & $0.853$ \\
\cline{3-4}
&  & $[0,1,0,1]$ & $0.820$ \\
\hline
\multirow{3}{*}{$\hat{c}_{\text{v}}$} & \multirow{3}{*}{$c_3 (100 \%)$}  & $[0,0,0,1]$ & $0.916$ \\
\cline{3-4}
&  & $[0,0,1,1]$ & $0.859$ \\
\cline{3-4}
&  & $[0,0,1,0]$ & $0.847$ \\
\hline
\multirow{3}{*}{$\hat{c}_{\text{vi}}$} & \multirow{3}{*}{$c_4 (100 \%)$}  & $[0,0,1,1]$ & $0.934$ \\
\cline{3-4}
&  & $[0,0,1,0]$ & $0.925$ \\
\cline{3-4}
&  & $[0,0,0,1]$ & $0.830$ \\
\hline
\end{tabular}
\end{center}
\end{table}

\subsection{Tests using Exploration--Exploitation Strategy}
\label{sec:exploration_exploitation_tets}

Here we evaluate the performance improvement obtained when using the exploration--exploitation strategy presented in Sec. \ref{sec:exploration_exploitation} w.r.t. the original PC search of Sec. \ref{sec:new_metrics}.
We proceed by testing the two approaches on a high-dimensional synthetic dataset.
A good estimator of the search efficiency is the mean CQ (MCQ) over all accepted clusters as a function of time (i.e., algorithm iterations). Of course, after a short initial transient a higher MCQ value, at every given time step, indicates a faster identification of the RPCs.
By definition, the MCQ ranges from $\tau_{\text{CQ}}$ to 1, and the maximum execution time (measured in number of iterations) is a user-defined setting.

The generated dataset lies in a 30-dimensional space and it is characterized by ten well-separated clusters.
The PC space consists of binary vectors of 30 parameters, so there are $2^{30}-1$ possible PCs (the all zeros configuration is never considered). We defined the Hamming distance as the dissimilarity measure $d_{\text{PC}} (\cdot, \cdot)$ used for comparing different PCs (see Sec. \ref{sec:exploration_exploitation}). Given a PC which is returned by an explorer, the exploiters generate similar PCs that have a hamming-distance equal to 1 from the original selected PC. This means that an exploiter randomly switches a parameter of the exploited PC to obtain the new candidate PC to be tested. 

\begin{figure}[ht!]
 \centering
 \includegraphics[viewport=0 0 844 474,scale=0.3,keepaspectratio=true]{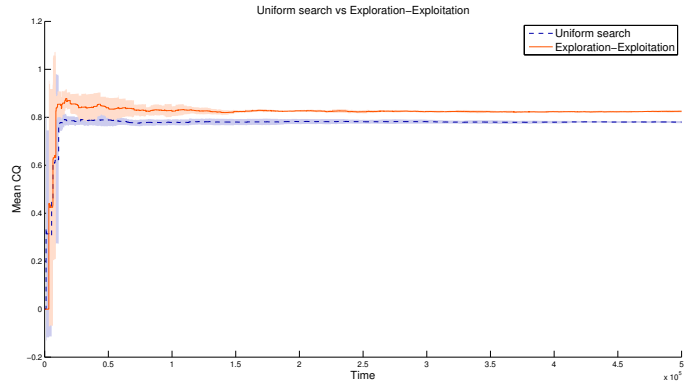}
 \caption{MCQ as a function of the time steps. The shaded areas represent the standard deviation of the measured CQs at each time stamp.}
\label{fig:mcqexplexpl}
\end{figure}

Fig. \ref{fig:mcqexplexpl} shows a plot of the MCQ obtained by both search methods, the uniform and the exploration--exploitation search, as a function of time.
The exploration-exploitation setting has been run with the ratio $\tau_{\text{EXPL}} = 3/4$ over a total of 4 agents, i.e., 3 explorers and 1 exploiter. Such results are intended as the average of five different runs considered for each method, executed by changing the random seeds.
As it is possible to observe, the MCQ obtained with the exploration--exploitation strategy rapidly assumes higher values w.r.t. those of the uniform search, and this behavior is preserved until convergence.

Please note that we are not reporting the results obtained by applying the exploration--exploitation strategy on the experiments described in the previous sections, since there are no significant variations that it is worth to discuss. In fact, since the dimension of the parameter space was reasonably small (we usually considered less than ten parameters), the basic version which explores the PCs with a uniform search was capable of considering a sufficient number of configurations for identifying the desired solution.
We must remark that in the asymptotic regime the results obtained with the two methods are the same, since all the PCs sooner or later will be considered. In this way, the tangible improvement introduced by the exploration--exploitation method consists in identifying the RPCs sooner, rather than discovering ``better solutions'' that cannot be found by the former technique.

%%%%%%%%
\section{Conclusions}
\label{sec:conclusions}

With this study we presented a dissimilarity-based multi-agent system, LD-ABCD, capable of discovering relevant clusters in a dataset, whose elements are grouped according to different and possibly equivalent configurations (instances of parameter values) of the dissimilarity measure.
Agents in LD-ABCD perform multiple and independent random walks. Accordingly, each agent discovers and takes decisions independently over one cluster at a time.
The multiple parameter configurations highlight the characteristics of patterns within the cluster that are considered to be discriminative, and represent the key for interpreting and characterizing semantically the regularities found in the dataset.
As a first step, we represented the entire dataset as a weighted graph. The identified clusters are subgraphs whose quality is evaluated as a function of their conductance normalized w.r.t. to the bounds of the graph conductance.
Guiding the evolution of our system with a cluster quality measure based on the conductance allowed us to define a powerful tool for evaluating the effectiveness of a given configuration of the parameters and to identify well-formed clusters, as outlined also by the tests performed on the UCI datasets for classification.
We presented two different approaches for searching the parameters characterizing the dissimilarity measure: (i) a basic one which consists in extracting configurations of the dissimilarity function parameters by means of a uniform distribution and (ii) an improved search strategy in which the solutions are further improved by searching in their neighborhood (the exploitation search strategy). In this second strategy, agents are divided into two main families: the explorers and the exploiters.
Our work highly relied on the celebrated Cheeger's inequality as reference to define suitable bounds for the definition of the cluster quality.
In this paper, we employed a very fast approach for computing an approximation of the minimum conductance of a graph, which is based on the numerical approximation of eigenvalues using the power method. This solution proved to be very useful and handy in our practical implementation.
The discussed experiments showed how LD-ABCD is capable of identifying the characterizing parameters of the dissimilarity measure, locally tailored for each single discovered cluster. Furthermore, when applied on the UCI datasets with a known class structure, the clusters returned by our algorithm contain elements belonging mostly to the same class.

Our future work will be focused on applying our system for clusters and knowledge discovery to larger datasets.
Accordingly, we will focus on the aspects related to scalability and parallelization, showing how our algorithm can work by distributing the computations over different cores and/or distinct workstations, each of which would access a suitable fraction of the entire dataset.
In fact, since LD-ABCD does not produce a partition of the data, it could also operate on a suitable subset of the entire dataset only.

%\clearpage
\appendices
\section{Graph Conductance and Related Approximation}
\label{sec:graph_conductance}

Given a graph $G=(\mathcal{V}, \mathcal{E})$, with $n=|\mathcal{V}|$, the conductance of a cut induced by the subset $\mathcal{S}\subset\mathcal{V}$ is defined as:
\begin{equation}
\label{eq:conductance}
 \phi(S) = \frac{\sum_{u \in S} \sum_{v \in \overline{S}} A(u, v)}{\min(A(S), A(\overline{S}))},
\end{equation}
where $\overline{S}=\mathcal{V}\setminus\mathcal{S}$ and $A(S) = \sum_{u,v \in S} A(u, v)$ is the number of edges in $\mathcal{S}$.
If the graph is weighted, then $A(u, v)$ contains the weight (i.e., the strength) of the edge among $u$ and $v$; if it is not weighted then $A(u, v)$ is equal to one if and only if there is an edge among $u$ and $v$.
While computing the conductance (\ref{eq:conductance}) of any subset $\mathcal{S}\subset\mathcal{V}$ is simple, computing the conductance of the graph $\Phi(G)$ consists in solving the following NP-Hard problem \cite{chung1994}:
\begin{equation}
\label{eq:graph_conductance}
 \Phi(G) = \min_{S\subset \mathcal{V}} \phi(S).
\end{equation}

Finding the global optimum is unfeasible even for small graphs. As a consequence, many approximation techniques have been proposed so far \cite{Leighton:1999:MMM:331524.331526, arora2009expander, madry2010fast, gkantsidis2003conductance, sarma2011estimating}.

Among the many techniques, spectral techniques \cite{chung1994} provide a very powerful approach.
Let $\mathbf{A}$ be the (weighted) adjacency matrix of $G$, and let $\mathbf{D}$ be diagonal matrix containing the vertex degrees:
\begin{equation}
 \mathbf{D} = \mathrm{diag}(d_1,..,d_n), \mathrm{where}\ d_i = \sum_{j=1}^{n} A(i,j).
\end{equation}

Let us define the transition matrix $\mathbf{M}$ as:
\begin{equation}
 \mathbf{M} = \mathbf{D}^{-1} \mathbf{A}.
\end{equation}

The matrix $\mathbf{M}$ is not always symmetric. Therefore, it does not always admit a spectral representation of the form $\mathbf{M}=\mathbf{U}\Lambda\mathbf{U}^T$, where $\Lambda$ is a diagonal matrix containing the $n$ eigenvalues and $\mathbf{U}$ is a matrix containing the corresponding eigenvectors.
Notwithstanding, $\mathbf{M}$ is conjugate to a symmetric matrix, $\mathbf{N}$, which is defined as follows:
\begin{equation}
\label{eq:symmetric_tm}
 \mathbf{N} = \mathbf{D}^{-1/2}\mathbf{A}\mathbf{D}^{-1/2} = \mathbf{D}^{1/2}\mathbf{M}\mathbf{D}^{-1/2}.
\end{equation}

$\mathbf{M}$ and $\mathbf{N}$ have the same eigenvalues and the eigenvectors are linearly correlated \cite{Lovasz1996,chung1994}. The eigenvalues of $\mathbf{N}$ satisfy the following relation:
\begin{equation}
\label{eq:autovalN}
 1 = \lambda_1 > \lambda_2 \geq ... \geq \lambda_n \geq -1.
\end{equation}

The celebrated Cheeger inequality \cite{Lovasz1996} establishes an important relation among the conductance of $G$ (\ref{eq:graph_conductance}) with $\lambda_2$:
\begin{equation}
\frac{\Phi(G)^2}{8} \leq 1 - \lambda_2 \leq \Phi(G),
\end{equation}
which can be rewritten as:
\begin{equation}
\label{eq:cheeger_ineq}
 1-\lambda_2 \leq \Phi(G) \leq \sqrt{8(1-\lambda_2)}.
\end{equation}

By using the fact that $\phi(S) \geq \Phi(G)$ for any $\mathcal{S}\subset\mathcal{V}$, Eq. \ref{eq:cheeger_ineq} can be used as a local reference for a specific graph.
According to Eq. \ref{eq:cheeger_ineq}, it is possible to define the lower and the upper bound of the graph conductance as
\begin{align}
\label{eq:approx_conduct}
\mathrm{lb}(\Phi(G)) &= 1-\lambda_2, \\
\nonumber\mathrm{ub}(\Phi(G)) &= \sqrt{8(1-\lambda_2)},
\end{align}
which can be used for evaluating how much the conductance of a cut $\phi(S)$ is close to the conductance of the whole graph, $\Phi(G)$.

To make use of the bounds of Eq. \ref{eq:approx_conduct}, we need to compute the $\lambda_2$ eigenvalue.
The QR-decomposition \cite{trefethen1997numerical} is the most straightforward numerical technique for this purpose, which is however characterized by a cubic computational complexity.
To overcome this drawback, we can use the power method described in \cite{trefethen1997numerical}, a fast algorithm that is able to compute in pseudo-linear time the largest eigenvalue and related eigenvector of a positive semi definite (PSD) matrix.
Notably, the computational complexity of the power method is $O((\mathcal{V} + |\mathcal{E}|) \frac{1}{\epsilon} \log{\frac{|\mathcal{V}|}{\epsilon}})$, where $\epsilon\geq0$ is the approximation used in computing $\lambda_2$.
Alg. \ref{alg:PM1} describes the pseudo-code of the power method. The algorithm starts by randomly initializing a vector, $\mathbf{x}_0 \in [-1, 1]^n$; it returns the vector $\mathbf{x}_t = \widetilde{\mathbf{M}}^t \mathbf{x}_0$, where $\widetilde{\mathbf{M}}$ is the PSD under analysis.
The following theorem is an important result for the convergence of the power method \cite{arora2008geometry,hoory2006expander}.
\begin{theorem}
 For every PSD matrix $\widetilde{\mathbf{M}}$, positive integer $t$, a parameter $\epsilon > 0$ and a vector $\mathbf{x}_0$ randomly picked with uniform probability $p$ in $[-1,1]^n$, with $p> \frac{3}{16}$ over the choice of $\mathbf{x}_0$, the power method outputs a vector $\mathbf{x}_t$ such that
 \begin{equation}
  \frac{\mathbf{x}_t^\intercal \widetilde{\mathbf{M}} \mathbf{x}_t}{\mathbf{x}_t^\intercal \mathbf{x}_t} \geq \lambda_1(1-\epsilon) \frac{1}{1+4n(1-\epsilon)^{2t}},
 \end{equation}
where $\lambda_1$ is the largest eigenvalue.
\end{theorem}

The eigenvector $\mathbf{v}_1$ related to $\lambda_1$ would be approximated by $\frac{\mathbf{x}_t}{\| \mathbf{x}_t \|}$.
Given a PSD matrix $\widetilde{\mathbf{M}}$ and the (unitary) eigenvector $\mathbf{v}_1$ related to $\lambda_1$, we can compute $\lambda_2$ by means of Alg. \ref{alg:PM2}, which is a variation of Alg. \ref{alg:PM1}.
The algorithm (\ref{alg:PM2}) returns a vector $\mathbf{x}_t \bot \mathbf{v}_1$, such that,
\begin{equation}
 \frac{\mathbf{x}_t^\intercal \widetilde{\mathbf{M}} \mathbf{x}_t}{\mathbf{x}_t^\intercal \mathbf{x}_t} \geq \lambda_2(1-\epsilon) \frac{1}{1+4n(1-\epsilon)^{2t}}.
\end{equation}

The power method can only be applied to a PSD matrix, which is not the case of $\mathbf{N}$, whose eigenvalues are the ones in Eq. \ref{eq:autovalN}. Consider now the matrix $\overline{\mathbf{N}} = \mathbf{N} + \mathbf{I}$. Every eigenvector of $\mathbf{N}$ with eigenvalue $\lambda$ is clearly also an eigenvector of $\overline{\mathbf{N}}$ with eigenvalue $1+\lambda$ and vice-versa, thus $\overline{\mathbf{N}}$ has eigenvalues $2 = 1+\lambda_1 > 1+\lambda_2 \geq ... \geq 1+\lambda_n \geq 0$ and thus it is PSD.

By using $\mathbf{v}_1$ (an eigenvector of $\lambda_1$ computed with Alg. \ref{alg:PM1}), and setting $t = O(\epsilon^{-1} \log{\frac{n}{\epsilon}})$, Alg. \ref{alg:PM2} will find with probability at least $3/16$ a vector $\mathbf{x}_t \bot \mathbf{1}$ such that
\begin{equation}
\label{eq:last}
 \frac{\mathbf{x}_t^\intercal \widetilde{\mathbf{M}} \mathbf{x}_t}{\mathbf{x}_t^\intercal \mathbf{x}_t} \geq \lambda_2 -4 \epsilon.
\end{equation}

From Eq. \ref{eq:last}, it is possible to derive the approximation of $\lambda_2$ that in turn can be used in Eq. \ref{eq:approx_conduct}.
\begin{algorithm}\footnotesize
\caption{Power method algorithm.}
\label{alg:PM1}
\begin{algorithmic}[1]
\REQUIRE PSD matrix $\widetilde{\mathbf{M}}$, tolerance $\epsilon$
\ENSURE Approximation of eigenvector $\mathbf{v}_1$ and related eigenvalue $\lambda_1$

\STATE{Pick random vector $\mathbf{x}_0 \in \{ 1, -1 \}^n$ with uniform probability;}

\STATE{$t = \epsilon^{-1} \log{\frac{n}{\epsilon}}$}
\FOR {i = 1 \TO t }
    \STATE{$\mathbf{x}_i = \widetilde{\mathbf{M}} \cdot \mathbf{x}_{i-1}$;}       
        \STATE{$\mathbf{x}_i = \frac{\mathbf{x}_i}{|| \mathbf{x}_i ||}$;}
\ENDFOR
\STATE $\mathbf{v}_1$ = $\mathbf{x}_t$
\STATE $\lambda_1 = \frac{\mathbf{x}_t^\intercal \widetilde{\mathbf{M}} \mathbf{x}_t}{\mathbf{x}_t^\intercal \mathbf{x}_t}$
\RETURN    $\mathbf{v}_1$, $\lambda_1$; 

\end{algorithmic}
\end{algorithm}
\begin{algorithm}\footnotesize
\caption{Computation of the second eigenvalue.}
\label{alg:PM2}
\begin{algorithmic}[1]
\REQUIRE PSD matrix $\widetilde{\mathbf{M}}$, eigenvector $\mathbf{v}_1$, and tolerance $\epsilon$
\ENSURE Approximation of $\lambda_2$

\STATE{Pick random vector $\mathbf{x}_0 \in \{ 1, -1 \}^n$ with uniform probability;}
   
\STATE{$\mathbf{x}_0 = \mathbf{x}_0 - \langle \mathbf{v}_1 \cdot \mathbf{x}_0 \rangle \mathbf{v}_1$;}
\STATE{$t = \epsilon^{-1} \log{\frac{n}{\epsilon}}$}
\FOR {i = 1 \TO t }
	\STATE{
        $\mathbf{x}_i = \widetilde{\mathbf{M}} \cdot \mathbf{x}_{i-1}$;}
        \STATE{
        $\mathbf{x}_i = \frac{\mathbf{x}_i}{|| \mathbf{x}_i ||}$;}
        \STATE{
        $\mathbf{x}_i = \mathbf{x}_i - \langle \mathbf{v}_1 \cdot \mathbf{x}_i \rangle \mathbf{v}_1$;}

\ENDFOR
\RETURN    $\lambda_2 = \frac{\mathbf{x}_t^\intercal \widetilde{\mathbf{M}} \mathbf{x}_t}{\mathbf{x}_t^\intercal \cdot \mathbf{x}_t}-1$; 

\end{algorithmic}
\end{algorithm}

\clearpage

\section*{Acknowledgements}
The work presented in this paper has been partially funded by Telecom Italia S.p.a.
The authors wish to thank Corrado Moiso, Software System Architect at Telecom Italia -- Future Centre, for the benefits that he provided to the present project trough the valuable comments, ideas and assistance to the writing and the undertaking of the research summarized here.

\bibliographystyle{abbrvnat}
\bibliography{Bibliography.bib}
\end{document}